\documentclass[runningheads,a4paper]{llncs}

\usepackage{amssymb, amsmath}
\usepackage{graphicx}
\usepackage{siunitx}
\usepackage{hyperref}
\usepackage{array}
\usepackage{color, colortbl}
\usepackage{nicefrac}
\usepackage{tikz}
\usepackage{hhline}
\usepackage{float}

\usepackage{pgfplots}
  \pgfplotsset{compat=newest}
  \usetikzlibrary{plotmarks}
  \usetikzlibrary{arrows.meta}
  \usepgfplotslibrary{patchplots}
  \usepackage{grffile}
  \usepackage{amsmath}

\newcommand{\etal}{et al.\ }
\renewcommand{\L}{\mathcal{L}}
\newcommand{\GDL}{\ensuremath{L_\text{GD}}}
\newcommand{\DL}{\ensuremath{L_\text{D}}}
\newcommand{\DGDL}{\ensuremath{L_\text{D or GD}}}
\newcommand{\CEL}{\ensuremath{L_\text{C}}}
\newcommand{\figref}[1]{Fig.~\ref{#1}}
\newcommand{\secref}[1]{Sec.~\ref{#1}}
\newcommand{\tabref}[1]{Tab.~\ref{#1}}

\newcommand{\tnhl}{\tabularnewline\hline}
\newcommand{\tn}{\tabularnewline}

\definecolor{gray}{gray}{0.9}

\begin{document}

\title{Spinal Cord Gray Matter-White Matter Segmentation on Magnetic Resonance AMIRA Images with MD-GRU}

\titlerunning{SC GM-WM Segmentation with MD-GRU}

\author{
Antal Horv\'{a}th\thanks{These two authors contributed equally.}\inst{1} \and 
Charidimos Tsagkas$^{\star}$\inst{2} \and 
Simon Andermatt\inst{1} \and
Simon Pezold\inst{1} \and 
Katrin Parmar\inst{2} \and 
Philippe Cattin\inst{1}}

\institute{
  \,\inst{1}Department of Biomedical Engineering, University of Basel, Allschwil, Switzerland\\
  \,\inst{2}Department of Neurology, University Hospital Basel, Basel, Switzerland
}

\authorrunning{A. Horv\'{a}th \etal}

\maketitle

\begin{abstract}
The small butterfly shaped structure of spinal cord (SC) gray matter (GM) is challenging to image and to delinate from its surrounding white matter (WM).
Segmenting GM is up to a point a trade-off between accuracy and precision.
We propose a new pipeline for GM-WM magnetic resonance (MR) image acquisition and segmentation.
We report superior results as compared to the ones recently reported in the SC GM segmentation challenge and show even better results using the averaged magnetization inversion recovery acquisitions (AMIRA) sequence.
Scan-rescan experiments with the AMIRA sequence show high reproducibility in terms of Dice coefficient, Hausdorff distance and relative standard deviation.
We use a recurrent neural network (RNN) with multi-dimensional gated recurrent units (MD-GRU) to train segmentation models on the AMIRA dataset of 855 slices.
We added a generalized dice loss to the cross entropy loss that MD-GRU uses and were able to improve the results.
\end{abstract}

\section{Introduction}

Cervical spinal cord (SC) segmentation in magnetic resonance (MR) images is a viable means for quantitatively assessing the neurodegenerative effects of diseases in the central nervous system.
While conventional MR sequences only allowed differentiation of the boundary between SC and cerebrospinal fluid (CSF), more recent sequences can be used to distinguish the SC's inner gray matter (GM) and white matter (WM) compartments. 
The latter task, however, remains challenging as state-of-the-art MR sequences only achieve an in-slice resolution of around 0.5 mm while maintaining a good signal-to-noise ratio (SNR) and an acceptable acquisition time.
This resolution is barely enough to visualize the SC's butterfly-shaped GM structure.

The 2016 spinal cord gray matter segmentation (SCGM) challenge \cite{prados_spinal_2017} reported mean Dice similarity coefficients (DSC) of 0.8 in comparison to a manual consensus ground truth for the best SC GM segmentation approaches at that time.
Porisky et al. \cite{porisky_grey_2017} experimented with 3D convolutional encoder networks but did not improve the challenge's results.
Perone et al.'s U-Net approach \cite{perone_spinal_2018} later managed to push the DSC value to 0.85.
More recently, Datta \etal \cite{datta_gray_2017} reported mean DSC of 0.88 on images of various MR sequences with a morphological geodesic active contour model.

Still, this means that a high number of subjects would be necessary to get reliable findings from clinical trials.
Hence, despite recent developments, there is a need for improvement of the reproducibility of SC GM and WM measurements.
An accurate and precise segmentation of the SC's inner structures in MR images under the mentioned limiting trade-off between resolution, SNR, and time therefore remains a challenge, especially when focusing on the GM.

In this work, we present a new robust and fully automatic pipeline for the acquisition and segmentation of GM and WM in MR images of the SC.
On the segmentation side, we propose the use of multi-dimensional gated recurrent units (MD-GRU), which already proved fit for a number of medical segmentation tasks \cite{andermatt2016multi},
to gain accurate and precise SC GM and WM segmentations.
To this end, we adapt MD-GRU's original cross-entropy loss by integrating a generalized Dice loss (GDL) \cite{sudre_generalised_2017} and show improved segmentation performance compared to the original.
Using the proposed setup, we manage to set a new state of the art on the SCGM challenge data with a mean DSC of 0.9.
On the imaging side, we propose to use the AMIRA MR sequence \cite{weigel_spinal_2018} for gaining improved GM-WM and WM-CSF contrast in axial cross-sectional slices of the SC.
Using the proposed MD-GRU approach in combination with this new imaging sequence, we manage to gain an even higher accuracy of DSC 0.91 wrt. a manual ground truth,
as we demonstrate in experiments on scan-rescan images of healthy subjects, for both SC GM and WM.

The remaining paper is structured as follows: in \secref{seq:method}, we present our segmentation method;
in \secref{seq:data}, we briefly describe the AMIRA MR sequence and the two datasets (SCGM challenge, AMIRA images) that we use for the experiments of \secref{seq:ExpResults}, before we conclude in \secref{seq:conclusion}.

\section{Method}
\label{seq:method}
The Multi-Dimensional Gated Recurrent Unit (MD-GRU) \cite{andermatt2016multi} is a generalization of a bi-directional recurrent neural network (RNN), which is able to process images.
It achieves this task by treating each direction along each of the spatial dimensions independently as a temporal direction. 
The MD-GRU processes the image using two convolutional GRUs (C-GRUs) for each image dimension, one in forward and one in backward direction, and combines the results of all individual C-GRUs. 
The gated recurrent unit (GRU), compared to the more popular and established long short-term memory (LSTM), uses a simpler gating structure and combines its state and output. 
The GRU has been shown to produce comparable results while consuming less memory than its LSTM counterpart when applied to image segmentation and hence allows for larger images to be processed \cite{andermatt2016multi}.

We directly feed the 2D version of MD-GRU the 8-channel AMIRA images (cf. \secref{seq:AMIRAimages}) to train AMIRA segmentation models, 
but only use the single channel images of the SCGM dataset (cf. \secref{seq:GMchallengeImages}) for the challenge models.
To address the high class imbalance between background, WM and GM, similar to \cite{perone_spinal_2018} we added a GM Dice loss (DL), 
but also included DLs for all the other label classes using the generalized Dice loss (GDL) formulation of Sudre \etal \cite{sudre_generalised_2017}.

\subsection{Dice Loss}
A straightforward approximation of a DL for a multi-labelling problem is
\begin{equation}
 \label{DiceLoss}
 \DL = - \, \frac{1}{\sum_{l\in\L}\omega_l}\,\sum\limits_{l\in\L} \omega_l\, \frac{2\,\sum_{x\in X} p_{lx}\,r_{lx}}{\sum_{x\in X}p_{lx}+r_{lx}},
\end{equation}
with the image domain $X$, labels $\L$, predictions $p$, raters $r$, and class weights $\omega$.
Sudre \etal \cite{sudre_generalised_2017} described a Generalized Dice Loss (GDL) $\GDL$ where they divide the weighted sum of the intersections of all labels by the weighted sum of all predictions and targets of all labels, 
instead of just linearly combining the individual Dice coefficients:
\begin{equation}
 \label{GeneralizedDiceLoss}
 \GDL = - \, \frac{2\,\sum_{l\in\L} \omega_l\sum_{x\in X} p_{lx}\,r_{lx}}{\sum_{l\in\L} \omega_l\sum_{x\in X}p_{lx}+r_{lx}}.
\end{equation}
As stated in \cite{crum_generalized_2006}, compared to the DL \eqref{DiceLoss}, the GDL \eqref{GeneralizedDiceLoss} allows all labels to contribute equally to the overall overlap (denominator in \eqref{GeneralizedDiceLoss}).

The (squared) inverse volume weighting
\begin{equation}
  \label{labelWeightsCalc}
  \omega_l=\frac{1}{\left(\sum_{x\in X} r_{lx}\right)^2},
\end{equation}
as proposed in \cite{crum_generalized_2006}, deals with the class imbalance problem: 
large regions only contribute very little to $\DL$ or $\GDL$, whereas small regions are weighted more and thus are more important in the optimization process.

To avoid division by zero in $\omega_l$ for image samples with absence of label $l$, we regularize the denominator of \eqref{labelWeightsCalc} and formulate the weighting we used:
\begin{equation}
  \label{labelWeightsCalcAdded}
  \omega_l=\frac{1}{1 + \left(\sum_{x\in X} r_{lx}\right)^2}.
\end{equation}
The weighting \eqref{labelWeightsCalcAdded} compared to \eqref{labelWeightsCalc} only slightly decreases its value as long as the object of interest has enough pixels.
Note, that during training of a network, it is possible, that not all labels occur in a random subsample with random location.

Finally, we combine DL or GDL with the cross entropy loss $\CEL$ (CEL) with a factor $\lambda\in[0,1]$:
$$L = \lambda \, \DGDL + (1-\lambda) \, \CEL.$$

\section{Data}
\label{seq:data}
In the following subsections, we describe the images used for the experiments: healthy subjects scan-rescan AMIRA dataset (own), which we call the AMIRA dataset,
and the SCGM challenge dataset\footnote{\url{http://cmictig.cs.ucl.ac.uk/niftyweb/program.php?p=CHALLENGE} {last accessed: \today} } \cite{prados_spinal_2017}, which we refer to as SCGM dataset.

\subsection{AMIRA Dataset}
\label{seq:AMIRAimages}

\begin{figure}[t]

  \begin{tabular}{c}
  \resizebox{\textwidth}{!}{
    \includegraphics[height=0.15\textwidth]{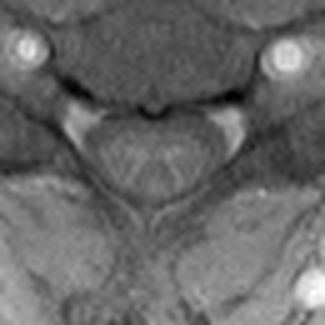}\,
    \includegraphics[height=0.15\textwidth]{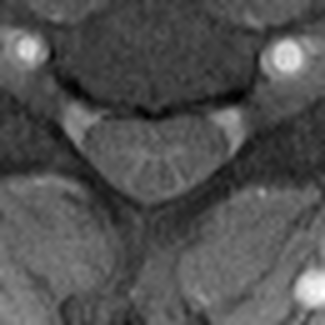}\,
    \includegraphics[height=0.15\textwidth]{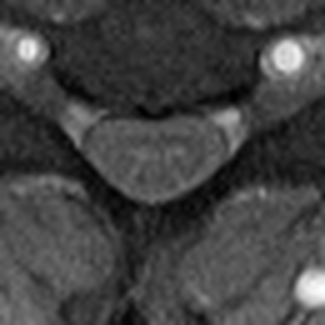}\,
    \includegraphics[height=0.15\textwidth]{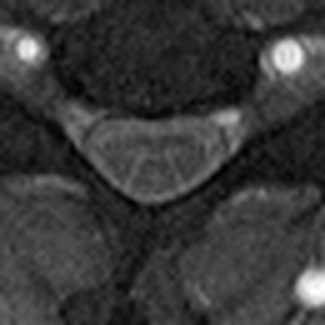}\,
    \includegraphics[height=0.15\textwidth]{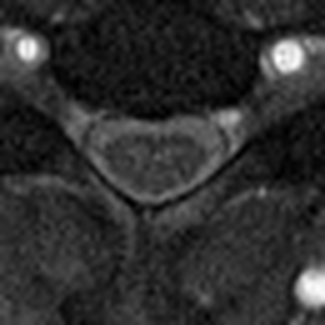}\,
    \includegraphics[height=0.15\textwidth]{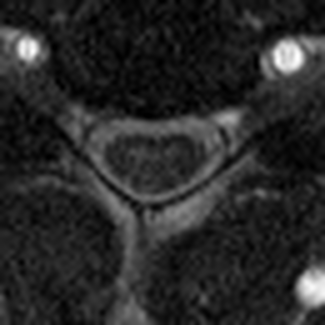}\,
    \includegraphics[height=0.15\textwidth]{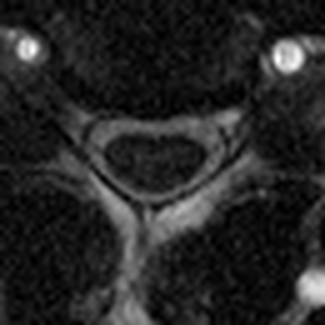}\,
    \includegraphics[height=0.15\textwidth]{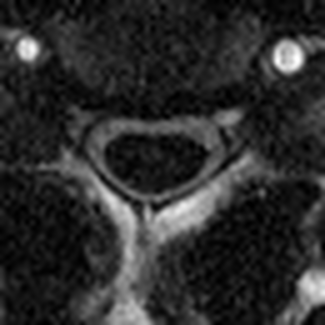}
  }\\
  \resizebox{\textwidth}{!}{
    \includegraphics[height=0.15\textwidth]{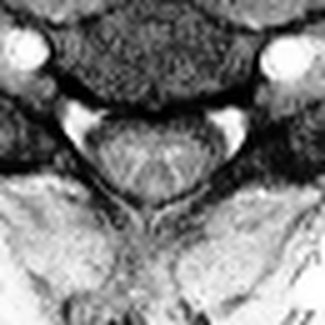}\,
    \includegraphics[height=0.15\textwidth]{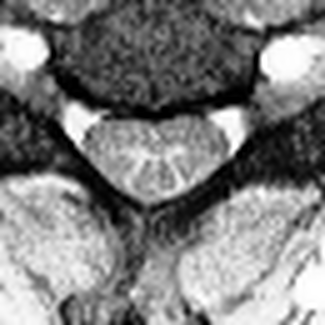}\,
    \includegraphics[height=0.15\textwidth]{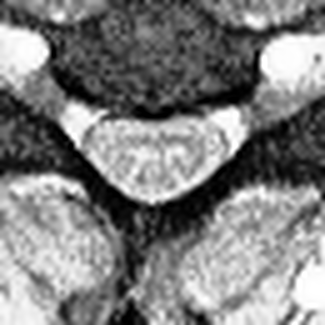}\,
    \includegraphics[height=0.15\textwidth]{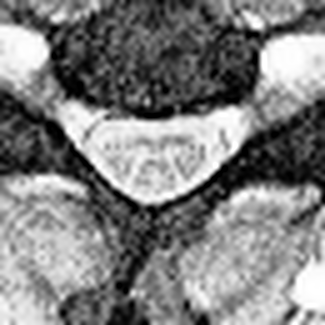}\,
    \includegraphics[height=0.15\textwidth]{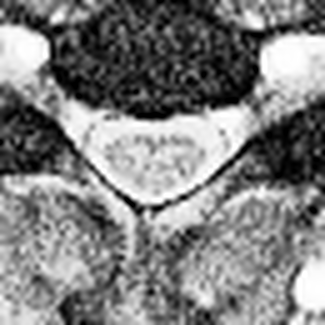}\,
    \includegraphics[height=0.15\textwidth]{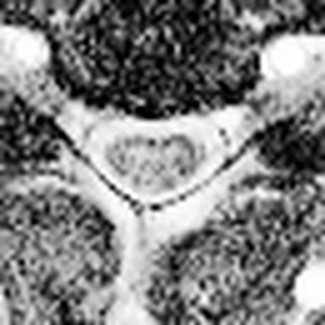}\,
    \includegraphics[height=0.15\textwidth]{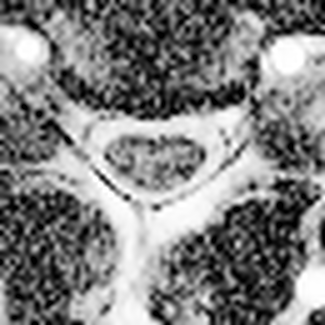}\,
    \includegraphics[height=0.15\textwidth]{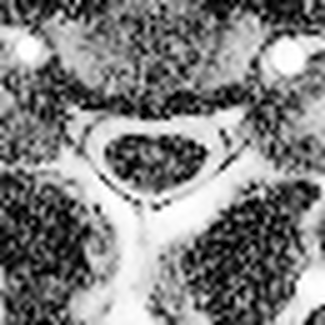}
  }\\
  \resizebox{\textwidth}{!}{
    \includegraphics[height=0.33\textwidth, trim = 10 10 10 10, clip]{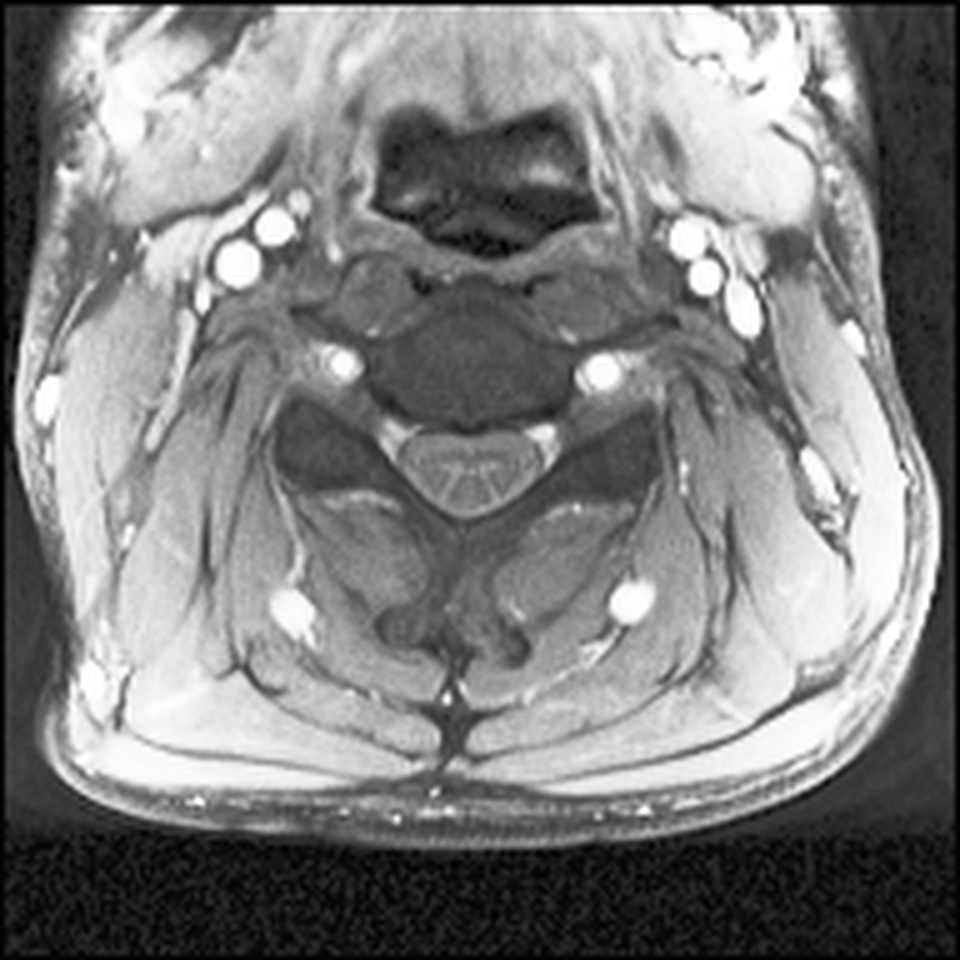}\,
    \includegraphics[height=0.33\textwidth]{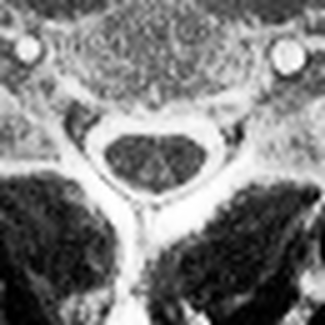}\,
    \includegraphics[height=0.33\textwidth]{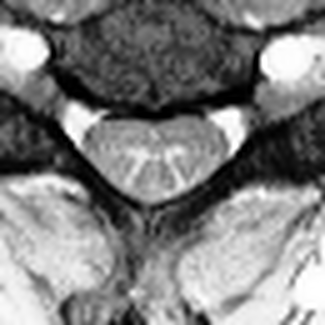}
  }

  \end{tabular}
  
  \caption{
  AMIRA sequence of an exemplary slice on C4 level. All images 10-fold upsampled.
  \emph{Top and middle row:} Inversion images with increasing inversion times from left to right.
  Original cropped images (\emph{top}), and histogram equalized (\emph{middle}).
  \emph{Bottom row:} Histogram equalized sum of the first 5 inversion images in full view (\emph{left}), weighted average with optimal CSF-WM contrast (\emph{middle}), and optimal GM-WM contrast (\emph{right}).
  }
  \label{fig:AMIRA}
\end{figure}

The first dataset used in this paper consists of 24 healthy subjects (14 female, 10 male, age $40\pm11$ years). 
Each subject was scanned 3 times, remaining in the scanner between the first and second scan, and leaving the scanner and being repositioned between the second and third scan.
Each scan contains 12 axial cross-sectional slices of the neck acquired with the AMIRA sequence \cite{weigel_spinal_2018} that were manually aligned at acquisition time perpendicular to the SC's centerline with an
average slice distance of $\SI{4}{\milli\meter}$ starting from vertebra C3 level in caudal direction.

Because of severe imaging artifacts some slices had to be discarded.
For one scan the last three caudal slices, for two scans the last two slices and for another two scans the last slice,
in total 9 out of the 864 slices were discarded.

The AMIRA sequence consists of 8 inversion images of the same anatomical slice captured at different inversion times after 180 degree MR pulses 
that have an in-slice resolution of $\SI{0.67}{\milli\meter}\times\SI{0.67}{\milli\meter}$.
Exemplary inversion images and different averages of an exemplary slice on vertebra C4 level are shown in \figref{fig:AMIRA}.
For human raters, to manually segment the AMIRA images, different single channel projections of the 8 channel images are necessary.
Weighted averages of the inversion images with e.g. optimal CSF-WM or GM-WM contrast, see \figref{fig:AMIRA}, were calculated with an approach that maximizes between-class intensity mean values and minimizes within-class intensity variances \cite{horvath_average_2018}.

In order to reduce the numerical errors for the calculated measures, we 10-fold upsampled all slices with Lanczos interpolation.
Since all images were manually centered at the SC, we consequently trimmed one third of the image size on each side and thus cropped out the inner ninth to a size of $650\times 650$ pixels for faster processing.

One experienced rater segmented all 855 images manually for WM and GM and segmented again 60 randomly chosen slices over all subjects, scans and slices, without knowledge of their origin, to enable an intra-rater comparison.

\subsection{SCGM Dataset}
\label{seq:GMchallengeImages}
The SCGM segmentation challenge data \cite{prados_spinal_2017} consists of 40 training datasets and 40 test datasets acquired at 4 different sites.
Both training and test datasets each have 10 samples of each site.
The 4 sites have different imaging protocols with different field of view, size and resolution.
Each dataset was manually segmented by 4 experts and to assess rater performance, with majority voting (more than 2 positive votes) a consensus segmentation of the 4 raters was calculated.

For training and testing of our MD-GRU models, we resampled all axial slices of all the datasets to the common finest resolution of $\SI{0.25}{\milli\meter}\times\SI{0.25}{\milli\meter}$
and center cropped or padded all datasets to a common size of $640\times 640$ pixels.
Before submitting the testing results for evaluation, we padded and resampled  all slices to their original sizes and resolutions.

\section{Experiments and Results}
\label{seq:ExpResults}
In the following subsections, we describe our experiments, the chosen MD-GRU options, and show their results.

\subsection{AMIRA segmentation model}
We split the 24 subjects into 3 groups of 8 subjects each for 3 cross-validations: training on two groups and testing on a third group.
To handle over-fitting, of each training set we excluded one subject and used it for validation.

We used the standard MD-GRU\footnote{\url{https://github.com/zubata88/mdgru} last accessed: \today} model with default settings and
residual learning, dropout rate 0.5, and dropconnect on state.
We chose the following problem specific parameters: Gaussian high pass filtering with variance 10, batch size 1, and window size $500\times 500$ pixels.
In each iteration of the training stage, for data augmentation, a subsample of the training data with random deformation field at a random location was selected.
Random deformations included an interpolated deformation field on 4 supporting points with randomly generated deformations of standard deviation of 15,
random scaling of a factor between $\nicefrac{4}{5}$ and $\nicefrac{5}{4}$, random rotation of $\pm$ 10 degrees, and random mirroring along the anatomical median plane.
To prevent zero padding of the subsamples, we only allowed random sampling within a safe distance of 45 pixels from the image boundary 
and truncated the random deformation magnitudes to 45 pixels, which is 3 times the chosen standard deviation.

\begin{figure}
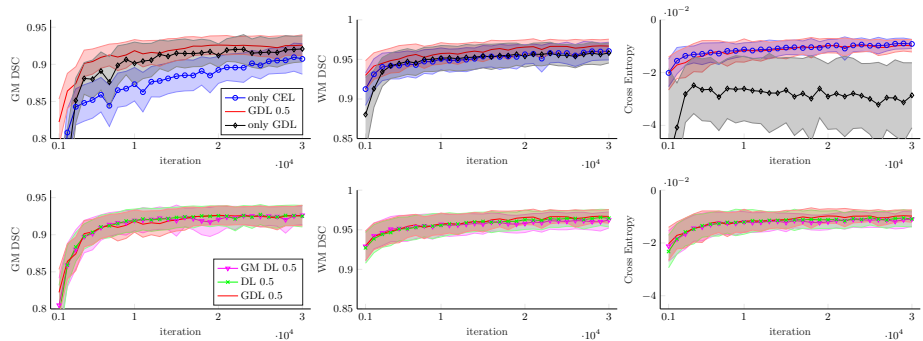

  \begin{tabular}{c}
    \resizebox{\textwidth}{!}{
    \newlength\fheight 
    \newlength\fwidth 
    \setlength\fheight{3.5cm} 
    \setlength\fwidth{8cm}
    \input{figs/errBars/GMDSCCELvsGDL.tex}
    \input{figs/errBars/WMDSCCELvsGDL.tex}
    \input{figs/errBars/crossEntropyCELvsGDL.tex}
    }\\
    \resizebox{\textwidth}{!}{
    \setlength\fheight{3.5cm} 
    \setlength\fwidth{8cm}
    \input{figs/errBars/GMDSCGDLvsDLvsGMDL.tex}
    \input{figs/errBars/WMDSCGDLvsDLvsGMDL.tex}
    \input{figs/errBars/crossEntropyGDLvsDLvsGMDL.tex}
    }
  \end{tabular}
    
 \caption{GM DSC, WM DSC and cross entropy over the training iterations of the validation set of group 1 in the AMIRA dataset in the format mean $\pm$ one standard deviation.
 \emph{Top row:} models with $\lambda=0$ (only CEL), $\lambda=1$ (only GDL), and combined with $\lambda=0.5$ (GDL 0.5).
 \emph{Bottom row:} GM DL 0.5, DL 0.5, and GDL 0.5 show similar performance.
 }
 \label{fig:errBars}
\end{figure}

We trained the networks with Adadelta with a learning rate of 1 for 30'000 iterations, where one iteration approximately took 10 seconds on an NVIDIA GeForce GTX Titan X.
Cross entropy and DSC on the evaluation set already reached their upper bounds after around 20'000 iterations, and dropconnect on state prevented from overfitting as we can see in \figref{fig:errBars}.

The time for segmenting a slice with the trained network approximately took 7 seconds.

Prior to the final model generation, we experimented in adding only a GM DL to the CEL with weightings $\lambda=0, 0.25, 0.5, 0.75, 1$ 
and figured that 0.5 produced the best results.
DL produces values close to -1 whereas CEL tends to have small values close to 0.
Moreover, CEL holds the information of all labels, since it is calculated over all labels.
Now, when adding only GM DL, because of the imbalance of the loss values, higher values of $\lambda$ strongly weaken the information for WM and background that in this setup is carried only within CEL.
The best weighting $\lambda$ depends on the cross entropy and thus depends on the class imbalance and label uncertainty of each specific segmentation task.

\begin{figure}

  \begin{tikzpicture}
  \node[anchor=north west,inner sep=0] at (0,0) {
    \begin{tabular}{c}
    \resizebox{\textwidth}{!}{
      \includegraphics[height=0.15\textwidth]{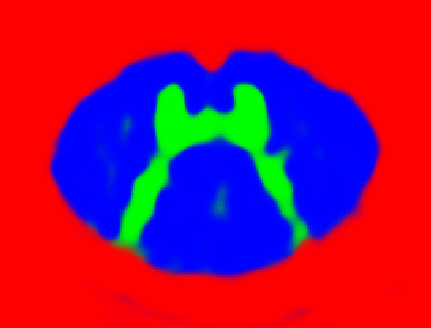}
      \includegraphics[height=0.15\textwidth]{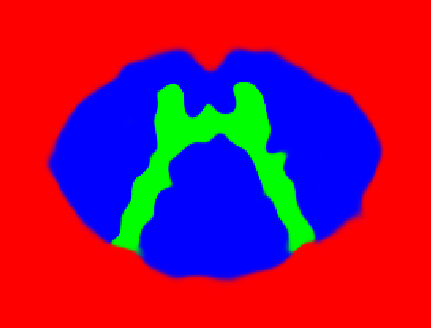}
      \includegraphics[height=0.15\textwidth]{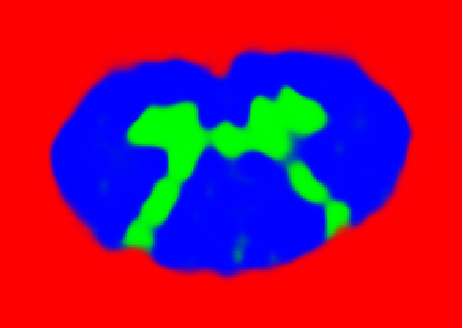}
      \includegraphics[height=0.15\textwidth]{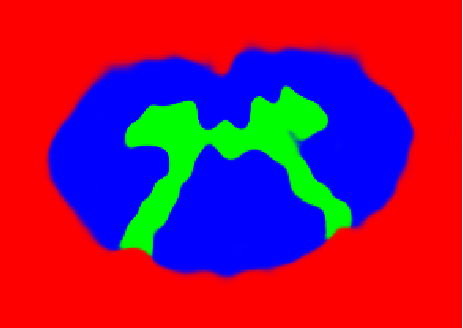}
    }
    \end{tabular}
  };

  \def\y{0.15}

  \node[] at (1.4,\y) {\tiny CGM 1 Scan 1 Slice 2};
  \node[] at (7.25,\y) {\tiny CGM 1 Scan 3 Slice 11};

  \def\y{-2.4}

  \node[] at (0.45,\y) {\tiny CEL};
  \node[] at (3.6,\y) {\tiny GDL 0.5};
  \node[] at (6.35,\y) {\tiny CEL};
  \node[] at (9.6,\y) {\tiny GDL 0.5};

  \end{tikzpicture}

 \caption{Exemplary prediction probability maps of the three labeling maps background (\emph{red}), GM (\emph{green}) and WM (\emph{blue}) of MD-GRU with CEL and with GDL in RGB colors.}
 \label{fig:CE_GDL_probabilitymaps}
\end{figure}
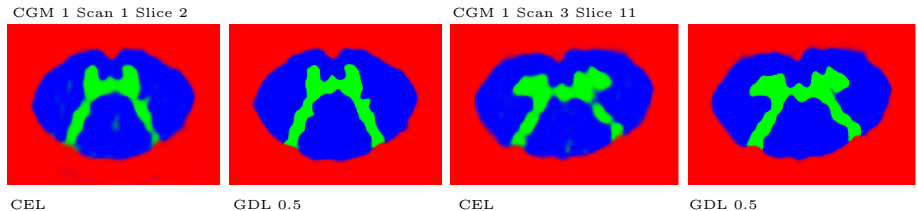

We observed that the auxiliary DL produces sharper probability maps at the boundaries as compared to only using CEL, see \figref{fig:CE_GDL_probabilitymaps},
and that DL helps to delineate weak contrasts e.g. between GM and WM.

Further experiments showed, that the proposed automatic weightings $\omega_l$ \eqref{labelWeightsCalcAdded} for the DLs between all label classes is a good strategy
to simplify the selection of $\lambda$.
In our case, the evaluation scores did not show big differences for $\lambda$ in a range from 0.25 to 0.75, when using the class weights $\omega_l$ according to \eqref{labelWeightsCalcAdded} for both DL and GDL.
MD-GRU with the trivial linear combinations $\lambda=0$ (only CEL) and $\lambda=1$ (only GDL) did not perform as good as true combinations between the two losses.
We show the improvement in the scores of GDL with $\lambda=0.5$ in \figref{fig:errBars} and \tabref{tab:CE_GDL}.

\begin{table}
 \caption{Improvement between native MD-GRU with CEL and the proposed MD-GRU with GDL together with the manual segmentation's precision and intra-rater accuracy values.
 Intra-rater accuracy of the human expert was calculated for the 60 randomly chosen slices.
 }
 \label{tab:CE_GDL}
 \resizebox{\textwidth}{!}{%
  \begin{tabular}{l|rlrl|rlrlrlrlrlrl}
    \rowcolor{gray}
    	&\multicolumn{4}{l|}{\bf Accuracy} & \multicolumn{6}{l}{\bf Intra-session} & \multicolumn{6}{l}{\bf Inter-session} \tn
    \rowcolor{gray}
    \bf GM &\multicolumn{2}{l}{\bf DSC} &\multicolumn{2}{l|}{\bf HD(mm)} &\multicolumn{2}{l}{\bf DSC} &\multicolumn{2}{l}{\bf HD(mm)} &\multicolumn{2}{l}{\bf RSD(\%)} &\multicolumn{2}{l}{\bf DSC} &\multicolumn{2}{l}{\bf HD(mm)} &\multicolumn{2}{l}{\bf RSD(\%)} \tnhl
    \bf MD-GRU CEL & 0.90 & $\pm$ 0.04 & 0.68 & $\pm$ 0.43 & 0.89 & $\pm$ 0.03 & 0.71& $\pm$ 0.46 & 3.22 & $\pm$ 2.87 & 0.88 & $\pm$ 0.04 & 0.70 & $\pm$ 0.43 & 3.65 & $\pm$ 3.97 \tn
    \rowcolor{gray}
    \bf MD-GRU GDL 0.5 & 0.91 & $\pm$ 0.03 & 0.56 & $\pm$ 0.33 & 0.88 & $\pm$ 0.03 & 0.58 & $\pm$ 0.32 & 2.93 & $\pm$ 2.63 & 0.88 & $\pm$ 0.03 & 0.61 & $\pm$ 0.35 & 3.86 & $\pm$ 3.49 \tn
    \bf Manual 	       &      &		   &	  &	       & 0.86 & $\pm$ 0.03 & 0.67 & $\pm$ 0.24 & 5.55 & $\pm$ 4.11 & 0.85 & $\pm$ 0.03 & 0.71 & $\pm$ 0.27 & 6.27 & $\pm$ 4.70 \tn
    \rowcolor{gray}
    \bf Intra-rater    & 0.85 & $\pm$ 0.07 & 0.62 & $\pm$ 0.30 &&&&&&&&&&&& \vspace{4mm}\tn
    
    \rowcolor{gray}
    \bf WM &\multicolumn{2}{l}{\bf DSC} &\multicolumn{2}{l|}{\bf HD(mm)} &\multicolumn{2}{l}{\bf DSC} &\multicolumn{2}{l}{\bf HD(mm)} &\multicolumn{2}{l}{\bf RSD(\%)} &\multicolumn{2}{l}{\bf DSC} &\multicolumn{2}{l}{\bf HD(mm)} &\multicolumn{2}{l}{\bf RSD(\%)}\tnhl
    \bf MD-GRU CEL & 0.94 & $\pm$ 0.03& 0.47 & $\pm$ 0.26& 0.94 & $\pm$ 0.02& 0.51 & $\pm$ 0.25 & 2.07 & $\pm$ 2.16& 0.94 & $\pm$ 0.02 & 0.52 & $\pm$ 0.22 & 2.40 & $\pm$ 2.22 \tn
    \rowcolor{gray}
    \bf MD-GRU GDL 0.5 & 0.95 & $\pm$ 0.02 & 0.43 & $\pm$ 0.22 & 0.94 & $\pm$ 0.02 & 0.51 & $\pm$ 0.22 & 2.14 & $\pm$ 2.35 & 0.94 & $\pm$ 0.02 & 0.53 & $\pm$ 0.23 & 2.69 & $\pm$ 2.54 \tn
    \bf Manual 	       &      &		   &	  &	       & 0.93 & $\pm$ 0.02 & 0.54 & $\pm$ 0.13 & 3.78 & $\pm$ 3.32 & 0.92 & $\pm$ 0.02 & 0.58 & $\pm$ 0.15 & 4.59 & $\pm$ 3.77 \tn
    \rowcolor{gray}
    \bf Intra-rater    & 0.96 & $\pm$ 0.02 & 0.44 & $\pm$ 0.15 &&&&&&&&&&&& \tn
  \end{tabular}
 }
\end{table}

Finally, comparisons between GM DL 0.5, auto-weighted DL 0.5 and GDL 0.5, all with $\lambda=0.5$, are shown in \figref{fig:errBars} on the bottom row.
As can be expected, the similarity of the terms DL \eqref{DiceLoss} and GDL \eqref{GeneralizedDiceLoss} is reflected in their almost identical segmentation performance.

GM DL 0.5 shows comparable WM segmentation performance to the losses that have WM DL included.
This can be explained, because the GM boundary is part of the WM boundaries and thus influences the WM scores, and
furthermore the outer WM boundary is already well delineated even without any DL through the good CSF-WM contrast.
Choosing a DL as a surrogate for GM DSC only, as proposed in \cite{perone_spinal_2018}, is thus justifiable.

\begin{figure}
    \begin{tikzpicture}
    \node[anchor=north west,inner sep=0] at (0,0) {
      \begin{tabular}{c}
      \resizebox{\textwidth}{!}{
	\includegraphics[height=0.15\textwidth]{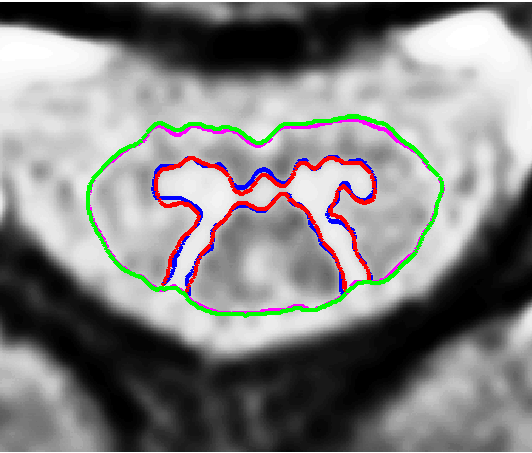}
	\includegraphics[height=0.15\textwidth]{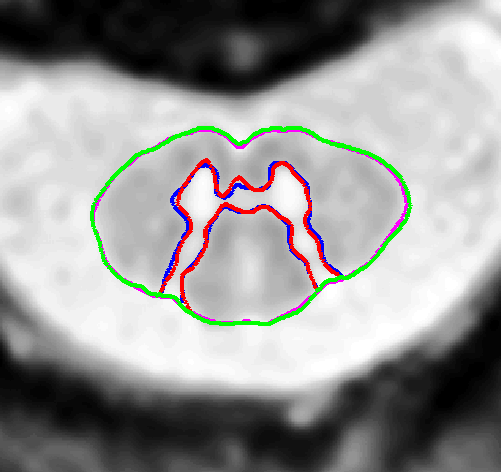}
	\includegraphics[height=0.15\textwidth]{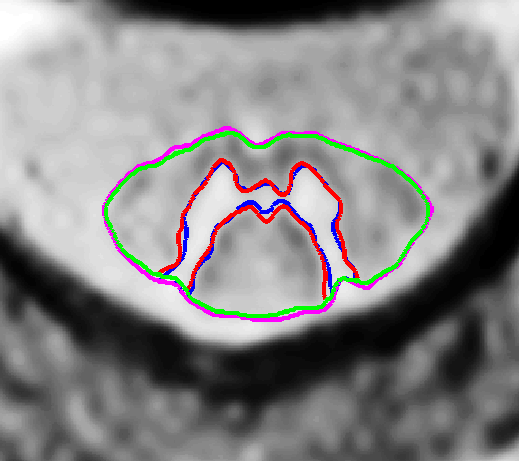}
	\includegraphics[height=0.15\textwidth]{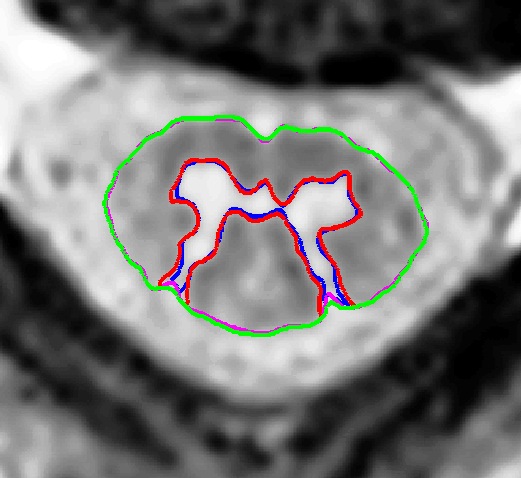}
	\includegraphics[height=0.15\textwidth]{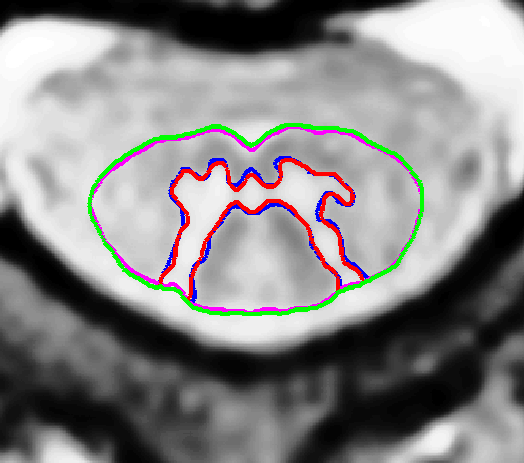}
	\includegraphics[height=0.15\textwidth]{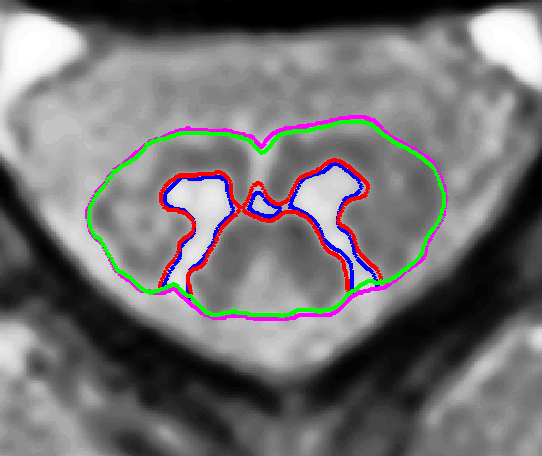}
      }
      \end{tabular}
    };

    \def\y{0.1}

    \node[] at (1.6,\y) {\tiny Subject 6537 Scan 1 Slice 10};
    \node[] at (5,\y) {\tiny Subject 6582 Scan 2 Slice 1};
    \node[] at (9,\y) {\tiny Subject 6537 Scan 1 Slice 5};
    
    \def\y{-1.9}

    \node[] at (3,\y) {\tiny Subject 6550 Scan 1 Slice 3};
    \node[] at (7,\y) {\tiny Subject 6614 Scan 1 Slice 7};
    \node[] at (10.7,\y) {\tiny Subject 6582 Scan 1 Slice 9};

    \end{tikzpicture}
  
  \caption{
    Exemplary slices of the AMIRA dataset with automatic GM (\emph{red}) and CSF-WM (\emph{green}) boundaries, and manual GM (\emph{blue}) and CSF-WM (\emph{magenta}) boundaries.
  }
  \label{fig:AMIRAresults}
\end{figure}

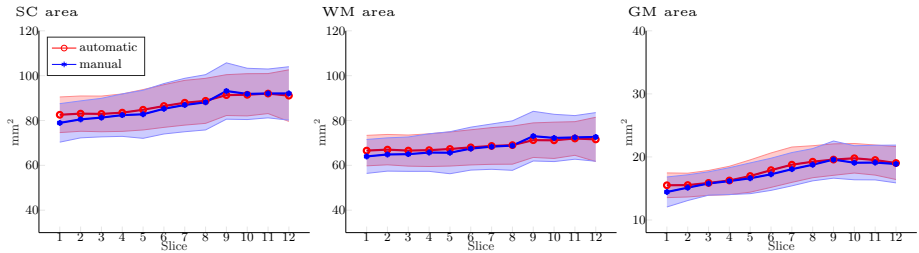
\begin{figure}
  \begin{tikzpicture}
    \node[anchor=north west,inner sep=0] at (0,0) {
	\resizebox{\textwidth}{!}{
	  \setlength\fheight{5cm} 
	  \setlength\fwidth{7cm}
%
%
\begin{tikzpicture}

\begin{axis}[%
width=0.951\fwidth,
height=\fheight,
at={(0\fwidth,0\fheight)},
scale only axis,
xmin=0,
xmax=13,
xtick={ 1,  2,  3,  4,  5,  6,  7,  8,  9, 10, 11, 12},
xlabel style={font=\color{white!15!black}},
xlabel={Slice},
ymin=30,
ymax=120,
ylabel style={font=\color{white!15!black}},
ylabel={$\text{mm}^\text{2}$},
axis background/.style={fill=white},
axis x line*=bottom,
axis y line*=left,
legend style={at={(0.03,0.97)}, anchor=north west, legend cell align=left, align=left, draw=white!15!black}
]

\addplot[area legend, draw=none, fill=red, fill opacity=0.2, forget plot]
table[row sep=crcr] {%
x	y\\
1	74.6071221507458\\
2	75.2020770081452\\
3	74.9894050865266\\
4	75.1491331502755\\
5	75.7856062964765\\
6	76.9390229234019\\
7	77.9502160950257\\
8	78.7284655187761\\
9	82.2234533657615\\
10	82.0573113558717\\
11	83.0966450778702\\
12	79.5886477232706\\
12	102.591669211849\\
11	100.918317158478\\
10	100.872034285929\\
9	100.431873594131\\
8	98.8029103431171\\
7	97.8942460897483\\
6	95.9572907946702\\
5	93.7887317456382\\
4	91.8434761870835\\
3	90.8901177372719\\
2	90.9278161136261\\
1	90.4946226792632\\
}--cycle;
\addplot [color=white!55!red, forget plot]
  table[row sep=crcr]{%
1	74.6071221507458\\
2	75.2020770081452\\
3	74.9894050865266\\
4	75.1491331502755\\
5	75.7856062964765\\
6	76.9390229234019\\
7	77.9502160950257\\
8	78.7284655187761\\
9	82.2234533657615\\
10	82.0573113558717\\
11	83.0966450778702\\
12	79.5886477232706\\
};
\addplot [color=white!55!red, forget plot]
  table[row sep=crcr]{%
1	90.4946226792632\\
2	90.9278161136261\\
3	90.8901177372719\\
4	91.8434761870835\\
5	93.7887317456382\\
6	95.9572907946702\\
7	97.8942460897483\\
8	98.8029103431171\\
9	100.431873594131\\
10	100.872034285929\\
11	100.918317158478\\
12	102.591669211849\\
};
\addplot [color=red, mark=o, mark options={solid, red}]
  table[row sep=crcr]{%
1	82.5508724150045\\
2	83.0649465608856\\
3	82.9397614118993\\
4	83.4963046686795\\
5	84.7871690210574\\
6	86.4481568590361\\
7	87.922231092387\\
8	88.7656879309466\\
9	91.3276634799461\\
10	91.4646728209004\\
11	92.007481118174\\
12	91.0901584675599\\
};
\addlegendentry{automatic}

\addplot [color=red, line width=1.5pt, mark=o, mark options={solid, red}, forget plot]
  table[row sep=crcr]{%
1	82.5508724150045\\
2	83.0649465608856\\
3	82.9397614118993\\
4	83.4963046686795\\
5	84.7871690210574\\
6	86.4481568590361\\
7	87.922231092387\\
8	88.7656879309466\\
9	91.3276634799461\\
10	91.4646728209004\\
11	92.007481118174\\
12	91.0901584675599\\
};

\addplot[area legend, draw=none, fill=blue, fill opacity=0.2, forget plot]
table[row sep=crcr] {%
x	y\\
1	70.3436099595494\\
2	72.2650562716101\\
3	72.7140258999506\\
4	72.9270487530909\\
5	72.0525649822762\\
6	73.9947239055706\\
7	74.9950766615255\\
8	75.7945313168062\\
9	80.5950267527037\\
10	80.4264424345832\\
11	81.1892997260269\\
12	80.1552533438779\\
12	103.977435629836\\
11	102.947273101706\\
10	103.292136626837\\
9	105.703263780834\\
8	100.434622338693\\
7	98.8162990071051\\
6	96.4670218404405\\
5	93.4832543515896\\
4	91.8611159358212\\
3	89.8874720580101\\
2	88.749774733372\\
1	87.5141835451461\\
}--cycle;
\addplot [color=white!55!blue, forget plot]
  table[row sep=crcr]{%
1	70.3436099595494\\
2	72.2650562716101\\
3	72.7140258999506\\
4	72.9270487530909\\
5	72.0525649822762\\
6	73.9947239055706\\
7	74.9950766615255\\
8	75.7945313168062\\
9	80.5950267527037\\
10	80.4264424345832\\
11	81.1892997260269\\
12	80.1552533438779\\
};
\addplot [color=white!55!blue, forget plot]
  table[row sep=crcr]{%
1	87.5141835451461\\
2	88.749774733372\\
3	89.8874720580101\\
4	91.8611159358212\\
5	93.4832543515896\\
6	96.4670218404405\\
7	98.8162990071051\\
8	100.434622338693\\
9	105.703263780834\\
10	103.292136626837\\
11	102.947273101706\\
12	103.977435629836\\
};
\addplot [color=blue, mark=asterisk, mark options={solid, blue}]
  table[row sep=crcr]{%
1	78.9288967523478\\
2	80.507415502491\\
3	81.3007489789804\\
4	82.3940823444561\\
5	82.7679096669329\\
6	85.2308728730056\\
7	86.9056878343153\\
8	88.1145768277496\\
9	93.1491452667689\\
10	91.85928953071\\
11	92.0682864138664\\
12	92.0663444868568\\
};
\addlegendentry{manual}

\addplot [color=blue, line width=1.5pt, mark=asterisk, mark options={solid, blue}, forget plot]
  table[row sep=crcr]{%
1	78.9288967523478\\
2	80.507415502491\\
3	81.3007489789804\\
4	82.3940823444561\\
5	82.7679096669329\\
6	85.2308728730056\\
7	86.9056878343153\\
8	88.1145768277496\\
9	93.1491452667689\\
10	91.85928953071\\
11	92.0682864138664\\
12	92.0663444868568\\
};
\end{axis}
\end{tikzpicture}%
%
%
\begin{tikzpicture}

\begin{axis}[%
width=0.951\fwidth,
height=\fheight,
at={(0\fwidth,0\fheight)},
scale only axis,
xmin=0,
xmax=13,
xtick={ 1,  2,  3,  4,  5,  6,  7,  8,  9, 10, 11, 12},
xlabel style={font=\color{white!15!black}},
xlabel={Slice},
ymin=30,
ymax=120,
ylabel style={font=\color{white!15!black}},
ylabel={$\text{mm}^\text{2}$},
axis background/.style={fill=white},
axis x line*=bottom,
axis y line*=left
]

\addplot[area legend, draw=none, fill=red, fill opacity=0.2, forget plot]
table[row sep=crcr] {%
x	y\\
1	59.783010113839\\
2	60.244545565264\\
3	59.651083458272\\
4	59.484303445506\\
5	59.6926213513551\\
6	60.1968205188671\\
7	60.4693116376272\\
8	60.5252962588523\\
9	63.5363645775216\\
10	63.0756803409857\\
11	64.483914385626\\
12	61.7043859464108\\
12	81.4332736160342\\
11	79.4722998482664\\
10	79.2542245532539\\
9	78.9317978643064\\
8	77.4999028571308\\
7	76.8225541356955\\
6	75.8496130156267\\
5	74.8807256018386\\
4	74.0534878303731\\
3	73.50744851211\\
2	73.7781839243005\\
1	73.3641637026274\\
}--cycle;
\addplot [color=white!55!red, forget plot]
  table[row sep=crcr]{%
1	59.783010113839\\
2	60.244545565264\\
3	59.651083458272\\
4	59.484303445506\\
5	59.6926213513551\\
6	60.1968205188671\\
7	60.4693116376272\\
8	60.5252962588523\\
9	63.5363645775216\\
10	63.0756803409857\\
11	64.483914385626\\
12	61.7043859464108\\
};
\addplot [color=white!55!red, forget plot]
  table[row sep=crcr]{%
1	73.3641637026274\\
2	73.7781839243005\\
3	73.50744851211\\
4	74.0534878303731\\
5	74.8807256018386\\
6	75.8496130156267\\
7	76.8225541356955\\
8	77.4999028571308\\
9	78.9317978643064\\
10	79.2542245532539\\
11	79.4722998482664\\
12	81.4332736160342\\
};
\addplot [color=red, mark=o, mark options={solid, red}, forget plot]
  table[row sep=crcr]{%
1	66.5735869082332\\
2	67.0113647447822\\
3	66.579265985191\\
4	66.7688956379396\\
5	67.2866734765968\\
6	68.0232167672469\\
7	68.6459328866613\\
8	69.0125995579916\\
9	71.234081220914\\
10	71.1649524471198\\
11	71.9781071169462\\
12	71.5688297812225\\
};
\addplot [color=red, line width=1.5pt, mark=o, mark options={solid, red}, forget plot]
  table[row sep=crcr]{%
1	66.5735869082332\\
2	67.0113647447822\\
3	66.579265985191\\
4	66.7688956379396\\
5	67.2866734765968\\
6	68.0232167672469\\
7	68.6459328866613\\
8	69.0125995579916\\
9	71.234081220914\\
10	71.1649524471198\\
11	71.9781071169462\\
12	71.5688297812225\\
};

\addplot[area legend, draw=none, fill=blue, fill opacity=0.2, forget plot]
table[row sep=crcr] {%
x	y\\
1	56.4166670784972\\
2	57.4331024462552\\
3	57.3250560485669\\
4	57.3275583769142\\
5	56.2663906101806\\
6	57.886947523492\\
7	58.2431428524684\\
8	57.7868159016314\\
9	61.9480373757431\\
10	61.6807016130146\\
11	62.7477309058725\\
12	61.8340389572787\\
12	83.5139031084004\\
11	82.1849732256659\\
10	82.7605025714829\\
9	84.0983971955956\\
8	79.8438153541929\\
7	78.4423031874632\\
6	77.0241773548023\\
5	74.9963388347136\\
4	74.0882573017629\\
3	72.6438460703018\\
2	72.2637008094261\\
1	71.5329753479079\\
}--cycle;
\addplot [color=white!55!blue, forget plot]
  table[row sep=crcr]{%
1	56.4166670784972\\
2	57.4331024462552\\
3	57.3250560485669\\
4	57.3275583769142\\
5	56.2663906101806\\
6	57.886947523492\\
7	58.2431428524684\\
8	57.7868159016314\\
9	61.9480373757431\\
10	61.6807016130146\\
11	62.7477309058725\\
12	61.8340389572787\\
};
\addplot [color=white!55!blue, forget plot]
  table[row sep=crcr]{%
1	71.5329753479079\\
2	72.2637008094261\\
3	72.6438460703018\\
4	74.0882573017629\\
5	74.9963388347136\\
6	77.0241773548023\\
7	78.4423031874632\\
8	79.8438153541929\\
9	84.0983971955956\\
10	82.7605025714829\\
11	82.1849732256659\\
12	83.5139031084004\\
};
\addplot [color=blue, mark=asterisk, mark options={solid, blue}, forget plot]
  table[row sep=crcr]{%
1	63.9748212132025\\
2	64.8484016278406\\
3	64.9844510594343\\
4	65.7079078393385\\
5	65.6313647224471\\
6	67.4555624391472\\
7	68.3427230199658\\
8	68.8153156279121\\
9	73.0232172856693\\
10	72.2206020922488\\
11	72.4663520657692\\
12	72.6739710328395\\
};
\addplot [color=blue, line width=1.5pt, mark=asterisk, mark options={solid, blue}, forget plot]
  table[row sep=crcr]{%
1	63.9748212132025\\
2	64.8484016278406\\
3	64.9844510594343\\
4	65.7079078393385\\
5	65.6313647224471\\
6	67.4555624391472\\
7	68.3427230199658\\
8	68.8153156279121\\
9	73.0232172856693\\
10	72.2206020922488\\
11	72.4663520657692\\
12	72.6739710328395\\
};
\end{axis}
\end{tikzpicture}%
%
%
\begin{tikzpicture}

\begin{axis}[%
width=0.951\fwidth,
height=\fheight,
at={(0\fwidth,0\fheight)},
scale only axis,
xmin=0,
xmax=13,
xtick={ 1,  2,  3,  4,  5,  6,  7,  8,  9, 10, 11, 12},
xlabel style={font=\color{white!15!black}},
xlabel={Slice},
ymin=8,
ymax=40,
ylabel style={font=\color{white!15!black}},
ylabel={$\text{mm}^\text{2}$},
axis background/.style={fill=white},
axis x line*=bottom,
axis y line*=left
]

\addplot[area legend, draw=none, fill=red, fill opacity=0.2, forget plot]
table[row sep=crcr] {%
x	y\\
1	13.5472146713749\\
2	13.6527265231371\\
3	13.9068714347908\\
4	14.0115096304113\\
5	14.4163740593415\\
6	15.1749472251578\\
7	15.9701599770565\\
8	16.7083070856311\\
9	17.0889203215631\\
10	17.4488822903491\\
11	17.1225319210053\\
12	16.4235177414803\\
12	21.6410964150498\\
11	21.8932530129803\\
10	22.132467585313\\
9	22.071577428926\\
8	21.7776226448594\\
7	21.5695968299699\\
6	20.6319698926285\\
5	19.5342465569871\\
4	18.5178762438308\\
3	17.8491810509505\\
2	17.4188814559281\\
1	17.4656278559404\\
}--cycle;
\addplot [color=white!55!red, forget plot]
  table[row sep=crcr]{%
1	13.5472146713749\\
2	13.6527265231371\\
3	13.9068714347908\\
4	14.0115096304113\\
5	14.4163740593415\\
6	15.1749472251578\\
7	15.9701599770565\\
8	16.7083070856311\\
9	17.0889203215631\\
10	17.4488822903491\\
11	17.1225319210053\\
12	16.4235177414803\\
};
\addplot [color=white!55!red, forget plot]
  table[row sep=crcr]{%
1	17.4656278559404\\
2	17.4188814559281\\
3	17.8491810509505\\
4	18.5178762438308\\
5	19.5342465569871\\
6	20.6319698926285\\
7	21.5695968299699\\
8	21.7776226448594\\
9	22.071577428926\\
10	22.132467585313\\
11	21.8932530129803\\
12	21.6410964150498\\
};
\addplot [color=red, mark=o, mark options={solid, red}, forget plot]
  table[row sep=crcr]{%
1	15.5064212636576\\
2	15.5358039895326\\
3	15.8780262428706\\
4	16.264692937121\\
5	16.9753103081643\\
6	17.9034585588932\\
7	18.7698784035132\\
8	19.2429648652452\\
9	19.5802488752446\\
10	19.7906749378311\\
11	19.5078924669928\\
12	19.0323070782651\\
};
\addplot [color=red, line width=1.5pt, mark=o, mark options={solid, red}, forget plot]
  table[row sep=crcr]{%
1	15.5064212636576\\
2	15.5358039895326\\
3	15.8780262428706\\
4	16.264692937121\\
5	16.9753103081643\\
6	17.9034585588932\\
7	18.7698784035132\\
8	19.2429648652452\\
9	19.5802488752446\\
10	19.7906749378311\\
11	19.5078924669928\\
12	19.0323070782651\\
};

\addplot[area legend, draw=none, fill=blue, fill opacity=0.2, forget plot]
table[row sep=crcr] {%
x	y\\
1	12.0645104410654\\
2	13.0878339601596\\
3	13.9376538259257\\
4	14.0384271534336\\
5	14.179041879447\\
6	14.6991326502715\\
7	15.4145214861255\\
8	16.2087616488863\\
9	16.6354632400944\\
10	16.3809253064454\\
11	16.3628276422173\\
12	15.890616235895\\
12	21.874163736032\\
11	21.8490924849066\\
10	21.7792037620159\\
9	22.5250345883775\\
8	21.3186495313738\\
7	20.6941240905363\\
6	19.7951918151631\\
5	19.0535540810619\\
4	18.2944155838115\\
3	17.6346949919188\\
2	17.1714282582496\\
1	16.8219121442536\\
}--cycle;
\addplot [color=white!55!blue, forget plot]
  table[row sep=crcr]{%
1	12.0645104410654\\
2	13.0878339601596\\
3	13.9376538259257\\
4	14.0384271534336\\
5	14.179041879447\\
6	14.6991326502715\\
7	15.4145214861255\\
8	16.2087616488863\\
9	16.6354632400944\\
10	16.3809253064454\\
11	16.3628276422173\\
12	15.890616235895\\
};
\addplot [color=white!55!blue, forget plot]
  table[row sep=crcr]{%
1	16.8219121442536\\
2	17.1714282582496\\
3	17.6346949919188\\
4	18.2944155838115\\
5	19.0535540810619\\
6	19.7951918151631\\
7	20.6941240905363\\
8	21.3186495313738\\
9	22.5250345883775\\
10	21.7792037620159\\
11	21.8490924849066\\
12	21.874163736032\\
};
\addplot [color=blue, mark=asterisk, mark options={solid, blue}, forget plot]
  table[row sep=crcr]{%
1	14.4432112926595\\
2	15.1296311092046\\
3	15.7861744089223\\
4	16.1664213686225\\
5	16.6162979802545\\
6	17.2471622327173\\
7	18.0543227883309\\
8	18.7637055901301\\
9	19.580248914236\\
10	19.0800645342306\\
11	19.105960063562\\
12	18.8823899859635\\
};
\addplot [color=blue, line width=1.5pt, mark=asterisk, mark options={solid, blue}, forget plot]
  table[row sep=crcr]{%
1	14.4432112926595\\
2	15.1296311092046\\
3	15.7861744089223\\
4	16.1664213686225\\
5	16.6162979802545\\
6	17.2471622327173\\
7	18.0543227883309\\
8	18.7637055901301\\
9	19.580248914236\\
10	19.0800645342306\\
11	19.105960063562\\
12	18.8823899859635\\
};
\end{axis}
\end{tikzpicture}%
	}
    };

    \def\y{0.2}
    \node[] at (0.7,\y) {\tiny SC area};
    \node[] at (4.8,\y) {\tiny WM area};
    \node[] at (8.8,\y) {\tiny GM area};

  \end{tikzpicture}
 \caption{SC, WM, and GM areas of GDL 0.5 (automatic) and manual segmentations wrt. the anatomical slice positions in mean $\pm$ one standard deviation.
 }
 \label{fig:areaSlicePlots}
\end{figure}

While the SCGM challenge results only provide GM segmentation accuracy, for the AMIRA dataset we additionally also provide WM segmentation results.
For the statistics, we gathered all slice-wise test results of all cross-validations for the proposed method GDL 0.5 and compare it with those of CEL.
Pairwise two-tailed Hotelling's T-tests for GM accuracy in DSC and labelmap Hausdorff distance (HD) show, that the test results of the MD-GRU models trained on the different groups are not significantly different from each other ($p>0.3$ for both GDL and CEL).

\begin{figure}
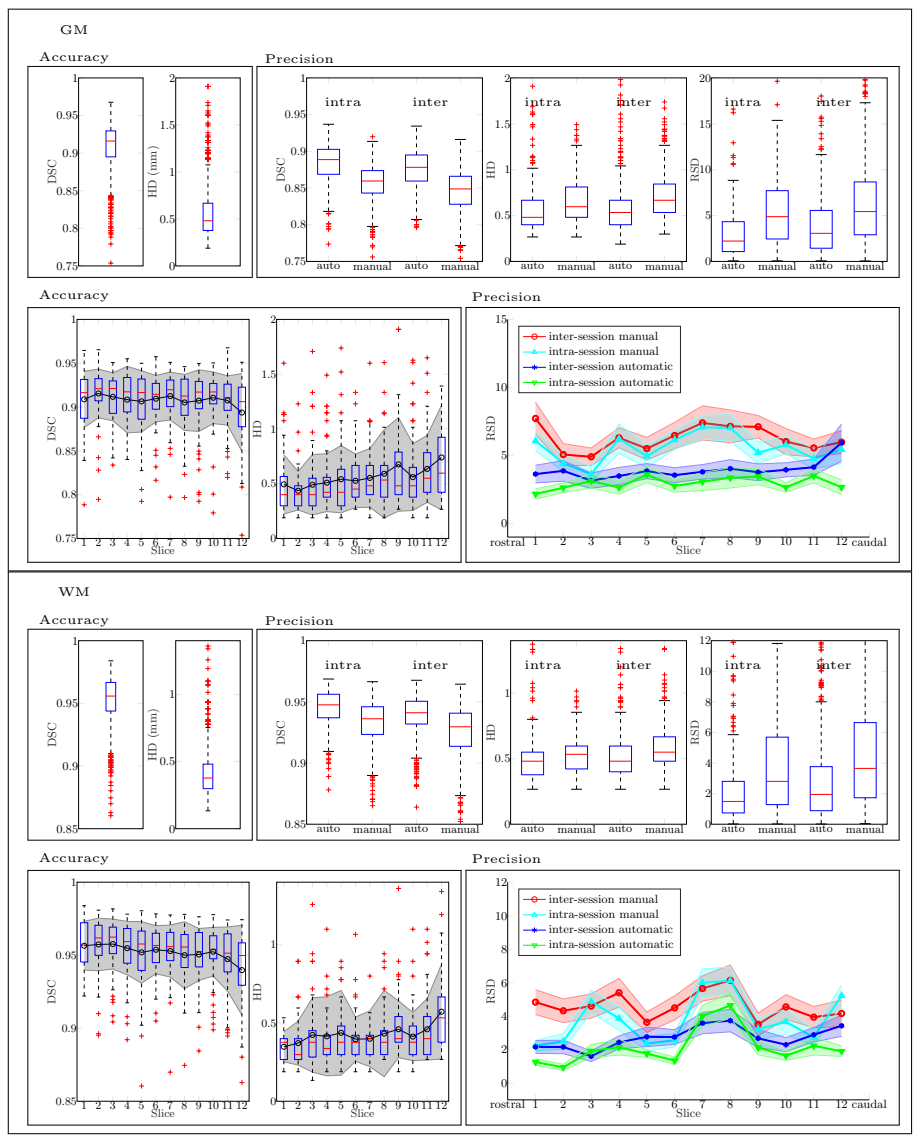


\resizebox{\textwidth}{!}{
\begin{tabular}{c}
\fbox{\begin{minipage}{\textwidth}
  \begin{tikzpicture}
    \node[anchor=north west,inner sep=0] at (0,0) {
      \begin{tabular}{c}
      \resizebox{\textwidth}{!}{
	\fbox{\resizebox{0.25\textwidth}{!}{
	  \setlength\fheight{5cm} 
	  \setlength\fwidth{1.79cm}
%
%
\begin{tikzpicture}

\begin{axis}[%
width=0.951\fwidth,
height=\fheight,
at={(0\fwidth,0\fheight)},
scale only axis,
xmin=0.5,
xmax=1.5,
xtick={\empty},
ymin=0.75,
ymax=1,
ylabel style={font=\color{white!15!black}},
ylabel={DSC},
axis background/.style={fill=white}
]
\addplot [color=black, dashed, forget plot]
  table[row sep=crcr]{%
1	0.929547246917121\\
1	0.967619047619048\\
};
\addplot [color=black, dashed, forget plot]
  table[row sep=crcr]{%
1	0.843251088534107\\
1	0.89502354367917\\
};
\addplot [color=black, forget plot]
  table[row sep=crcr]{%
0.9625	0.967619047619048\\
1.0375	0.967619047619048\\
};
\addplot [color=black, forget plot]
  table[row sep=crcr]{%
0.9625	0.843251088534107\\
1.0375	0.843251088534107\\
};
\addplot [color=blue, forget plot]
  table[row sep=crcr]{%
0.925	0.89502354367917\\
0.925	0.929547246917121\\
1.075	0.929547246917121\\
1.075	0.89502354367917\\
0.925	0.89502354367917\\
};
\addplot [color=red, forget plot]
  table[row sep=crcr]{%
0.925	0.916083916083916\\
1.075	0.916083916083916\\
};
\addplot [color=black, draw=none, mark=+, mark options={solid, red}, forget plot]
  table[row sep=crcr]{%
1	0.711111111111111\\
1	0.711673699015471\\
1	0.753709198813056\\
1	0.779150994089199\\
1	0.788448623237072\\
1	0.792375064399794\\
1	0.792383292383292\\
1	0.794636556104446\\
1	0.796756756756757\\
1	0.797153024911032\\
1	0.80038666022233\\
1	0.800637958532695\\
1	0.802598700649675\\
1	0.805890227576975\\
1	0.80865224625624\\
1	0.812741312741313\\
1	0.815513626834382\\
1	0.81640146878825\\
1	0.819654992158913\\
1	0.820903647250953\\
1	0.823170731707317\\
1	0.824069352371239\\
1	0.826913700633837\\
1	0.827438370846731\\
1	0.827936507936508\\
1	0.82837528604119\\
1	0.831533477321814\\
1	0.831672784478238\\
1	0.832451499118166\\
1	0.833759590792839\\
1	0.83729216152019\\
1	0.838896952104499\\
1	0.839087947882736\\
1	0.839132627332325\\
1	0.84009840098401\\
1	0.840594059405941\\
1	0.840712468193384\\
1	0.841296928327645\\
1	0.84185773074662\\
1	0.841904761904762\\
1	0.841926345609065\\
1	0.842240824211204\\
1	0.842527582748245\\
1	0.842938847338319\\
};
\end{axis}
\end{tikzpicture}%
%
%
\begin{tikzpicture}

\begin{axis}[%
width=0.951\fwidth,
height=\fheight,
at={(0\fwidth,0\fheight)},
scale only axis,
xmin=0.5,
xmax=1.5,
xtick={\empty},
ymin=0,
ymax=2,
ylabel style={font=\color{white!15!black}},
ylabel={HD (mm)},
axis background/.style={fill=white}
]
\addplot [color=black, dashed, forget plot]
  table[row sep=crcr]{%
1	0.666666701436043\\
1	1.07496775583707\\
};
\addplot [color=black, dashed, forget plot]
  table[row sep=crcr]{%
1	0.188561818150677\\
1	0.377123636301355\\
};
\addplot [color=black, forget plot]
  table[row sep=crcr]{%
0.9625	1.07496775583707\\
1.0375	1.07496775583707\\
};
\addplot [color=black, forget plot]
  table[row sep=crcr]{%
0.9625	0.188561818150677\\
1.0375	0.188561818150677\\
};
\addplot [color=blue, forget plot]
  table[row sep=crcr]{%
0.925	0.377123636301355\\
0.925	0.666666701436043\\
1.075	0.666666701436043\\
1.075	0.377123636301355\\
0.925	0.377123636301355\\
};
\addplot [color=red, forget plot]
  table[row sep=crcr]{%
0.925	0.480740195134419\\
1.075	0.480740195134419\\
};
\addplot [color=black, draw=none, mark=+, mark options={solid, red}, forget plot]
  table[row sep=crcr]{%
1	1.13137090890406\\
1	1.13137090890406\\
1	1.13137090890406\\
1	1.13920055878961\\
1	1.13920055878961\\
1	1.14697663390721\\
1	1.14697676209185\\
1	1.14697676209185\\
1	1.14697676209185\\
1	1.14697689027648\\
1	1.19256965019724\\
1	1.19256978347728\\
1	1.20000006258488\\
1	1.20000006258488\\
1	1.20000006258488\\
1	1.20738461311925\\
1	1.20738474805501\\
1	1.20738474805501\\
1	1.20738474805501\\
1	1.20738488299076\\
1	1.22927265841728\\
1	1.22927265841728\\
1	1.22927265841728\\
1	1.22927265841728\\
1	1.22927265841728\\
1	1.22927265841728\\
1	1.22927265841728\\
1	1.25786421654361\\
1	1.31318110872721\\
1	1.31318110872721\\
1	1.31318110872721\\
1	1.33333325386047\\
1	1.33333325386047\\
1	1.33333325386047\\
1	1.33333340287209\\
1	1.33333340287209\\
1	1.37275075705946\\
1	1.37275075705946\\
1	1.39204094045533\\
1	1.41735284895199\\
1	1.49071206274655\\
1	1.50849454520542\\
1	1.52023414931794\\
1	1.6000000834465\\
1	1.6000000834465\\
1	1.60554584814098\\
1	1.60554602757472\\
1	1.62754083364736\\
1	1.64924252057893\\
1	1.70749988570181\\
1	1.73845406538815\\
1	1.90904290800119\\
1	1.90904290800119\\
1	1.90904290800119\\
1	2.27058522096805\\
1	2.80317326502631\\
1	2.8503412588478\\
1	3.1467798231056\\
};
\end{axis}
\end{tikzpicture}%
	}}\,\fbox{\resizebox{0.75\textwidth}{!}{
	  \setlength\fheight{5cm} 
	  \setlength\fwidth{5cm}
%
%
\begin{tikzpicture}

\begin{axis}[%
width=0.951\fwidth,
height=\fheight,
at={(0\fwidth,0\fheight)},
scale only axis,
xmin=0.5,
xmax=4.5,
xtick={1,2,3,4},
xticklabels={{auto},{manual},{auto},{manual}},
ymin=0.75,
ymax=1,
ylabel style={font=\color{white!15!black}},
ylabel={DSC},
axis background/.style={fill=white}
]
\addplot [color=black, dashed, forget plot]
  table[row sep=crcr]{%
1	0.902680830412086\\
1	0.936829558998808\\
};
\addplot [color=black, dashed, forget plot]
  table[row sep=crcr]{%
2	0.873700801675865\\
2	0.913510457885811\\
};
\addplot [color=black, dashed, forget plot]
  table[row sep=crcr]{%
3	0.894957983193277\\
3	0.934322033898305\\
};
\addplot [color=black, dashed, forget plot]
  table[row sep=crcr]{%
4	0.865979381443299\\
4	0.916152897657213\\
};
\addplot [color=black, dashed, forget plot]
  table[row sep=crcr]{%
1	0.818143637127257\\
1	0.868696530192833\\
};
\addplot [color=black, dashed, forget plot]
  table[row sep=crcr]{%
2	0.797640653357532\\
2	0.843135418958034\\
};
\addplot [color=black, dashed, forget plot]
  table[row sep=crcr]{%
3	0.807106598984772\\
3	0.85948275862069\\
};
\addplot [color=black, dashed, forget plot]
  table[row sep=crcr]{%
4	0.771498771498772\\
4	0.827890556045896\\
};
\addplot [color=black, forget plot]
  table[row sep=crcr]{%
0.875	0.936829558998808\\
1.125	0.936829558998808\\
};
\addplot [color=black, forget plot]
  table[row sep=crcr]{%
1.875	0.913510457885811\\
2.125	0.913510457885811\\
};
\addplot [color=black, forget plot]
  table[row sep=crcr]{%
2.875	0.934322033898305\\
3.125	0.934322033898305\\
};
\addplot [color=black, forget plot]
  table[row sep=crcr]{%
3.875	0.916152897657213\\
4.125	0.916152897657213\\
};
\addplot [color=black, forget plot]
  table[row sep=crcr]{%
0.875	0.818143637127257\\
1.125	0.818143637127257\\
};
\addplot [color=black, forget plot]
  table[row sep=crcr]{%
1.875	0.797640653357532\\
2.125	0.797640653357532\\
};
\addplot [color=black, forget plot]
  table[row sep=crcr]{%
2.875	0.807106598984772\\
3.125	0.807106598984772\\
};
\addplot [color=black, forget plot]
  table[row sep=crcr]{%
3.875	0.771498771498772\\
4.125	0.771498771498772\\
};
\addplot [color=blue, forget plot]
  table[row sep=crcr]{%
0.75	0.868696530192833\\
0.75	0.902680830412086\\
1.25	0.902680830412086\\
1.25	0.868696530192833\\
0.75	0.868696530192833\\
};
\addplot [color=blue, forget plot]
  table[row sep=crcr]{%
1.75	0.843135418958034\\
1.75	0.873700801675865\\
2.25	0.873700801675865\\
2.25	0.843135418958034\\
1.75	0.843135418958034\\
};
\addplot [color=blue, forget plot]
  table[row sep=crcr]{%
2.75	0.85948275862069\\
2.75	0.894957983193277\\
3.25	0.894957983193277\\
3.25	0.85948275862069\\
2.75	0.85948275862069\\
};
\addplot [color=blue, forget plot]
  table[row sep=crcr]{%
3.75	0.827890556045896\\
3.75	0.865979381443299\\
4.25	0.865979381443299\\
4.25	0.827890556045896\\
3.75	0.827890556045896\\
};
\addplot [color=red, forget plot]
  table[row sep=crcr]{%
0.75	0.888702928870293\\
1.25	0.888702928870293\\
};
\addplot [color=red, forget plot]
  table[row sep=crcr]{%
1.75	0.859528487229863\\
2.25	0.859528487229863\\
};
\addplot [color=red, forget plot]
  table[row sep=crcr]{%
2.75	0.878169223953928\\
3.25	0.878169223953928\\
};
\addplot [color=red, forget plot]
  table[row sep=crcr]{%
3.75	0.848753251447023\\
4.25	0.848753251447023\\
};
\addplot [color=black, draw=none, mark=+, mark options={solid, red}, forget plot]
  table[row sep=crcr]{%
1	0.773509933774834\\
1	0.793931731984829\\
1	0.794463087248322\\
1	0.797385620915033\\
1	0.800884955752212\\
1	0.801526717557252\\
1	0.814814814814815\\
1	0.815189873417722\\
1	0.815454171804357\\
};
\addplot [color=black, draw=none, mark=+, mark options={solid, red}, forget plot]
  table[row sep=crcr]{%
2	0.728749323226854\\
2	0.75609756097561\\
2	0.769820971867008\\
2	0.771911298838437\\
2	0.782751540041068\\
2	0.784365393061045\\
2	0.786600496277916\\
2	0.789557805007991\\
2	0.790649350649351\\
2	0.791226645004062\\
2	0.793518903199003\\
2	0.796341463414634\\
2	0.919878296146045\\
};
\addplot [color=black, draw=none, mark=+, mark options={solid, red}, forget plot]
  table[row sep=crcr]{%
3	0.597427346355407\\
3	0.638953759925269\\
3	0.796037820801441\\
3	0.797758532857871\\
3	0.799829642248722\\
3	0.8\\
3	0.8055822906641\\
};
\addplot [color=black, draw=none, mark=+, mark options={solid, red}, forget plot]
  table[row sep=crcr]{%
4	0.7392138063279\\
4	0.74177897574124\\
4	0.744106300900129\\
4	0.748551564310545\\
4	0.754208754208754\\
4	0.765397273154678\\
4	0.765513454146074\\
4	0.76555023923445\\
4	0.769005847953216\\
4	0.770258236865539\\
};
\end{axis}
\end{tikzpicture}%
%
%
\begin{tikzpicture}

\begin{axis}[%
width=0.951\fwidth,
height=\fheight,
at={(0\fwidth,0\fheight)},
scale only axis,
xmin=0.5,
xmax=4.5,
xtick={1,2,3,4},
xticklabels={{auto},{manual},{auto},{manual}},
ymin=0,
ymax=2,
ylabel style={font=\color{white!15!black}},
ylabel={HD},
axis background/.style={fill=white}
]
\addplot [color=black, dashed, forget plot]
  table[row sep=crcr]{%
1	0.666666701436043\\
1	1.01543646707432\\
};
\addplot [color=black, dashed, forget plot]
  table[row sep=crcr]{%
2	0.811035023678478\\
2	1.2649111300376\\
};
\addplot [color=black, dashed, forget plot]
  table[row sep=crcr]{%
3	0.666666701436043\\
3	1.04136667776572\\
};
\addplot [color=black, dashed, forget plot]
  table[row sep=crcr]{%
4	0.843274086691736\\
4	1.2649111300376\\
};
\addplot [color=black, dashed, forget plot]
  table[row sep=crcr]{%
1	0.266666680574417\\
1	0.400000020861626\\
};
\addplot [color=black, dashed, forget plot]
  table[row sep=crcr]{%
2	0.266666680574417\\
2	0.480740195134419\\
};
\addplot [color=black, dashed, forget plot]
  table[row sep=crcr]{%
3	0.188561818150677\\
3	0.400000065565109\\
};
\addplot [color=black, dashed, forget plot]
  table[row sep=crcr]{%
4	0.29814241254931\\
4	0.533333361148834\\
};
\addplot [color=black, forget plot]
  table[row sep=crcr]{%
0.875	1.01543646707432\\
1.125	1.01543646707432\\
};
\addplot [color=black, forget plot]
  table[row sep=crcr]{%
1.875	1.2649111300376\\
2.125	1.2649111300376\\
};
\addplot [color=black, forget plot]
  table[row sep=crcr]{%
2.875	1.04136667776572\\
3.125	1.04136667776572\\
};
\addplot [color=black, forget plot]
  table[row sep=crcr]{%
3.875	1.2649111300376\\
4.125	1.2649111300376\\
};
\addplot [color=black, forget plot]
  table[row sep=crcr]{%
0.875	0.266666680574417\\
1.125	0.266666680574417\\
};
\addplot [color=black, forget plot]
  table[row sep=crcr]{%
1.875	0.266666680574417\\
2.125	0.266666680574417\\
};
\addplot [color=black, forget plot]
  table[row sep=crcr]{%
2.875	0.188561818150677\\
3.125	0.188561818150677\\
};
\addplot [color=black, forget plot]
  table[row sep=crcr]{%
3.875	0.29814241254931\\
4.125	0.29814241254931\\
};
\addplot [color=blue, forget plot]
  table[row sep=crcr]{%
0.75	0.400000020861626\\
0.75	0.666666701436043\\
1.25	0.666666701436043\\
1.25	0.400000020861626\\
0.75	0.400000020861626\\
};
\addplot [color=blue, forget plot]
  table[row sep=crcr]{%
1.75	0.480740195134419\\
1.75	0.811035023678478\\
2.25	0.811035023678478\\
2.25	0.480740195134419\\
1.75	0.480740195134419\\
};
\addplot [color=blue, forget plot]
  table[row sep=crcr]{%
2.75	0.400000065565109\\
2.75	0.666666701436043\\
3.25	0.666666701436043\\
3.25	0.400000065565109\\
2.75	0.400000065565109\\
};
\addplot [color=blue, forget plot]
  table[row sep=crcr]{%
3.75	0.533333361148834\\
3.75	0.843274086691736\\
4.25	0.843274086691736\\
4.25	0.533333361148834\\
3.75	0.533333361148834\\
};
\addplot [color=red, forget plot]
  table[row sep=crcr]{%
0.75	0.480740195134419\\
1.25	0.480740195134419\\
};
\addplot [color=red, forget plot]
  table[row sep=crcr]{%
1.75	0.596284825098619\\
2.25	0.596284825098619\\
};
\addplot [color=red, forget plot]
  table[row sep=crcr]{%
2.75	0.533333361148834\\
3.25	0.533333361148834\\
};
\addplot [color=red, forget plot]
  table[row sep=crcr]{%
3.75	0.666666701436043\\
4.25	0.666666701436043\\
};
\addplot [color=black, draw=none, mark=+, mark options={solid, red}, forget plot]
  table[row sep=crcr]{%
1	1.07496775583707\\
1	1.07496775583707\\
1	1.07496775583707\\
1	1.07496775583707\\
1	1.07496787597408\\
1	1.09949489084117\\
1	1.13137090890406\\
1	1.13920055878961\\
1	1.13920055878961\\
1	1.14697676209185\\
1	1.2649111300376\\
1	1.3333335518837\\
1	1.46666674315929\\
1	1.49071206274655\\
1	1.49071206274655\\
1	1.60000026226044\\
1	1.60554602757472\\
1	1.60554602757472\\
1	1.62754083364736\\
1	1.68654817338347\\
1	1.90904290800119\\
1	3.33333387970924\\
};
\addplot [color=black, draw=none, mark=+, mark options={solid, red}, forget plot]
  table[row sep=crcr]{%
2	1.31318110872721\\
2	1.33333340287209\\
2	1.37275075705946\\
2	1.37275075705946\\
2	1.37275091047631\\
2	1.43604402346473\\
2	1.4907122293466\\
};
\addplot [color=black, draw=none, mark=+, mark options={solid, red}, forget plot]
  table[row sep=crcr]{%
3	1.06666672229767\\
3	1.06666672229767\\
3	1.06666684150696\\
3	1.09949489084117\\
3	1.09949489084117\\
3	1.09949489084117\\
3	1.13137078246352\\
3	1.13137090890406\\
3	1.13137090890406\\
3	1.13137090890406\\
3	1.13920055878961\\
3	1.13920055878961\\
3	1.1925695169172\\
3	1.19256965019724\\
3	1.19256965019724\\
3	1.20000006258488\\
3	1.22927265841728\\
3	1.2649111300376\\
3	1.31318110872721\\
3	1.31993257954077\\
3	1.33998363578985\\
3	1.41735284895199\\
3	1.46666674315929\\
3	1.52023414931794\\
3	1.5549205863875\\
3	1.60554602757472\\
3	1.70749988570181\\
3	1.75372647918691\\
3	1.81352969842536\\
3	1.92296078053768\\
3	1.98214260301925\\
3	2.03087293414865\\
3	2.6195844972071\\
3	2.74873722710292\\
3	2.9242284457719\\
3	2.93333348631859\\
3	3.62705939685072\\
};
\addplot [color=black, draw=none, mark=+, mark options={solid, red}, forget plot]
  table[row sep=crcr]{%
4	1.31318110872721\\
4	1.31318110872721\\
4	1.31318110872721\\
4	1.31318125548663\\
4	1.31993272705474\\
4	1.33333340287209\\
4	1.33998333628022\\
4	1.35973860787397\\
4	1.37275075705946\\
4	1.37275075705946\\
4	1.37275075705946\\
4	1.41735284895199\\
4	1.41735284895199\\
4	1.43604386297431\\
4	1.43604386297431\\
4	1.44222058540326\\
4	1.4907122293466\\
4	1.52023397941856\\
4	1.5549205863875\\
4	1.67066196528403\\
4	1.73845406538815\\
4	2.2191092940194\\
};
\end{axis}
\end{tikzpicture}%
%
%
\begin{tikzpicture}

\begin{axis}[%
width=0.951\fwidth,
height=\fheight,
at={(0\fwidth,0\fheight)},
scale only axis,
xmin=0.5,
xmax=4.5,
xtick={1,2,3,4},
xticklabels={{auto},{manual},{auto},{manual}},
ymin=0,
ymax=20,
ylabel style={font=\color{white!15!black}},
ylabel={RSD},
axis background/.style={fill=white}
]
\addplot [color=black, dashed, forget plot]
  table[row sep=crcr]{%
1	4.33162585866399\\
1	8.83455445743969\\
};
\addplot [color=black, dashed, forget plot]
  table[row sep=crcr]{%
2	7.71151572828212\\
2	15.3707303319227\\
};
\addplot [color=black, dashed, forget plot]
  table[row sep=crcr]{%
3	5.55621340294841\\
3	11.6409125267641\\
};
\addplot [color=black, dashed, forget plot]
  table[row sep=crcr]{%
4	8.66353835036844\\
4	17.2902696207091\\
};
\addplot [color=black, dashed, forget plot]
  table[row sep=crcr]{%
1	0.0629378532431276\\
1	1.0754943765081\\
};
\addplot [color=black, dashed, forget plot]
  table[row sep=crcr]{%
2	0.0632192026094428\\
2	2.44673626708148\\
};
\addplot [color=black, dashed, forget plot]
  table[row sep=crcr]{%
3	0.0618911843489381\\
3	1.43926747772611\\
};
\addplot [color=black, dashed, forget plot]
  table[row sep=crcr]{%
4	0.0533866954463163\\
4	2.90591827884881\\
};
\addplot [color=black, forget plot]
  table[row sep=crcr]{%
0.875	8.83455445743969\\
1.125	8.83455445743969\\
};
\addplot [color=black, forget plot]
  table[row sep=crcr]{%
1.875	15.3707303319227\\
2.125	15.3707303319227\\
};
\addplot [color=black, forget plot]
  table[row sep=crcr]{%
2.875	11.6409125267641\\
3.125	11.6409125267641\\
};
\addplot [color=black, forget plot]
  table[row sep=crcr]{%
3.875	17.2902696207091\\
4.125	17.2902696207091\\
};
\addplot [color=black, forget plot]
  table[row sep=crcr]{%
0.875	0.0629378532431276\\
1.125	0.0629378532431276\\
};
\addplot [color=black, forget plot]
  table[row sep=crcr]{%
1.875	0.0632192026094428\\
2.125	0.0632192026094428\\
};
\addplot [color=black, forget plot]
  table[row sep=crcr]{%
2.875	0.0618911843489381\\
3.125	0.0618911843489381\\
};
\addplot [color=black, forget plot]
  table[row sep=crcr]{%
3.875	0.0533866954463163\\
4.125	0.0533866954463163\\
};
\addplot [color=blue, forget plot]
  table[row sep=crcr]{%
0.75	1.0754943765081\\
0.75	4.33162585866399\\
1.25	4.33162585866399\\
1.25	1.0754943765081\\
0.75	1.0754943765081\\
};
\addplot [color=blue, forget plot]
  table[row sep=crcr]{%
1.75	2.44673626708148\\
1.75	7.71151572828212\\
2.25	7.71151572828212\\
2.25	2.44673626708148\\
1.75	2.44673626708148\\
};
\addplot [color=blue, forget plot]
  table[row sep=crcr]{%
2.75	1.43926747772611\\
2.75	5.55621340294841\\
3.25	5.55621340294841\\
3.25	1.43926747772611\\
2.75	1.43926747772611\\
};
\addplot [color=blue, forget plot]
  table[row sep=crcr]{%
3.75	2.90591827884881\\
3.75	8.66353835036844\\
4.25	8.66353835036844\\
4.25	2.90591827884881\\
3.75	2.90591827884881\\
};
\addplot [color=red, forget plot]
  table[row sep=crcr]{%
0.75	2.2200659573621\\
1.25	2.2200659573621\\
};
\addplot [color=red, forget plot]
  table[row sep=crcr]{%
1.75	4.87861704365145\\
2.25	4.87861704365145\\
};
\addplot [color=red, forget plot]
  table[row sep=crcr]{%
2.75	3.06238101076364\\
3.25	3.06238101076364\\
};
\addplot [color=red, forget plot]
  table[row sep=crcr]{%
3.75	5.43607316654828\\
4.25	5.43607316654828\\
};
\addplot [color=black, draw=none, mark=+, mark options={solid, red}, forget plot]
  table[row sep=crcr]{%
1	10.574750060988\\
1	10.6509808044836\\
1	11.320466986738\\
1	11.5285257489655\\
1	12.9293798402643\\
1	16.2071013256254\\
1	16.5987507320786\\
};
\addplot [color=black, draw=none, mark=+, mark options={solid, red}, forget plot]
  table[row sep=crcr]{%
2	17.0798739417041\\
2	19.6347255030021\\
2	20.0059479555218\\
2	25.0649233005963\\
};
\addplot [color=black, draw=none, mark=+, mark options={solid, red}, forget plot]
  table[row sep=crcr]{%
3	12.1899159945315\\
3	12.2850875115239\\
3	12.5156993335285\\
3	13.37655897783\\
3	13.9118076819764\\
3	14.8184655917654\\
3	15.2612974356809\\
3	15.5744266456034\\
3	15.7289500539527\\
3	17.4395247587977\\
3	18.0232064633944\\
3	28.3651219513613\\
3	30.5829462577647\\
};
\addplot [color=black, draw=none, mark=+, mark options={solid, red}, forget plot]
  table[row sep=crcr]{%
4	17.9851072606144\\
4	18.271876908812\\
4	18.2865559448582\\
4	18.3406819883\\
4	18.4098806936143\\
4	18.5460183258482\\
4	19.0979267397392\\
4	19.2333044482741\\
4	19.752425220612\\
4	19.820734138523\\
4	20.0644958913628\\
4	20.1329014087836\\
4	20.899076216393\\
4	21.355072800354\\
4	21.4083976494591\\
4	22.2933054435188\\
4	23.1860516937487\\
4	23.4704930661825\\
4	24.3208701809386\\
};
\end{axis}
\end{tikzpicture}%
	}}
      }\\[4mm]
      \fbox{\resizebox{0.47\textwidth}{!}{
	\setlength\fheight{6.1cm} 
	\setlength\fwidth{5cm}
	\input{figs/MDGRU_plots/GM_55000_accDSC_slice.tex}
	\input{figs/MDGRU_plots/GM_55000_accHD_slice.tex}
      }}\,\fbox{\resizebox{0.47\textwidth}{!}{
	\setlength\fheight{5.765cm} 
	\setlength\fwidth{10cm}
%
%
\definecolor{mycolor1}{rgb}{0.00000,1.00000,1.00000}%
\begin{tikzpicture}

\begin{axis}[%
width=0.951\fwidth,
height=\fheight,
at={(0\fwidth,0\fheight)},
scale only axis,
xmin=0,
xmax=13,
xtick={0,1,2,3,4,5,6,7,8,9,10,11,12,13},
xticklabels={{rostral},{1},{2},{3},{4},{5},{6},{7},{8},{9},{10},{11},{12},{caudal}},
xlabel style={font=\color{white!15!black}},
xlabel={Slice},
ymin=-1,
ymax=15,
ylabel style={font=\color{white!15!black}},
ylabel={RSD},
axis background/.style={fill=white},
axis x line*=bottom,
axis y line*=left,
legend style={at={(0.03,0.97)}, anchor=north west, legend cell align=left, align=left, draw=white!15!black}
]

\addplot[area legend, draw=none, fill=red, fill opacity=0.2, forget plot]
table[row sep=crcr] {%
x	y\\
1	6.53565501347707\\
2	4.26639925945275\\
3	4.25727291114943\\
4	5.56487243579163\\
5	4.67584525205231\\
6	5.50382410025811\\
7	6.14014182194788\\
8	5.93473771596275\\
9	6.28360581179368\\
10	5.05606047761545\\
11	4.89562283651049\\
12	5.13902745170948\\
12	6.82055212342244\\
11	6.20992360054479\\
10	6.9752099729264\\
9	7.9332223152773\\
8	8.34644006722628\\
7	8.64636716613568\\
6	7.43439049694544\\
5	6.33185142393249\\
4	7.03681473633289\\
3	5.54588752301606\\
2	5.87107093817111\\
1	8.90566717753758\\
}--cycle;
\addplot [color=white!55!red, forget plot]
  table[row sep=crcr]{%
1	6.53565501347707\\
2	4.26639925945275\\
3	4.25727291114943\\
4	5.56487243579163\\
5	4.67584525205231\\
6	5.50382410025811\\
7	6.14014182194788\\
8	5.93473771596275\\
9	6.28360581179368\\
10	5.05606047761545\\
11	4.89562283651049\\
12	5.13902745170948\\
};
\addplot [color=white!55!red, forget plot]
  table[row sep=crcr]{%
1	8.90566717753758\\
2	5.87107093817111\\
3	5.54588752301606\\
4	7.03681473633289\\
5	6.33185142393249\\
6	7.43439049694544\\
7	8.64636716613568\\
8	8.34644006722628\\
9	7.9332223152773\\
10	6.9752099729264\\
11	6.20992360054479\\
12	6.82055212342244\\
};
\addplot [color=red, mark=o, mark options={solid, red}]
  table[row sep=crcr]{%
1	7.72066109550733\\
2	5.06873509881193\\
3	4.90158021708274\\
4	6.30084358606226\\
5	5.5038483379924\\
6	6.46910729860177\\
7	7.39325449404178\\
8	7.14058889159452\\
9	7.10841406353549\\
10	6.01563522527093\\
11	5.55277321852764\\
12	5.97978978756596\\
};
\addlegendentry{inter-session manual}

\addplot [color=red, line width=1.5pt, mark=o, mark options={solid, red}, forget plot]
  table[row sep=crcr]{%
1	7.72066109550733\\
2	5.06873509881193\\
3	4.90158021708274\\
4	6.30084358606226\\
5	5.5038483379924\\
6	6.46910729860177\\
7	7.39325449404178\\
8	7.14058889159452\\
9	7.10841406353549\\
10	6.01563522527093\\
11	5.55277321852764\\
12	5.97978978756596\\
};

\addplot[area legend, draw=none, fill=mycolor1, fill opacity=0.2, forget plot]
table[row sep=crcr] {%
x	y\\
1	5.35186719527366\\
2	3.81430664684546\\
3	3.06648210018055\\
4	5.15890305184559\\
5	4.11788551600075\\
6	5.45209252069895\\
7	6.198042243057\\
8	6.04647066566451\\
9	4.4040130782629\\
10	5.13534041074813\\
11	4.01745200958745\\
12	4.39966058634739\\
12	6.44962198499411\\
11	5.52786786427714\\
10	6.3974943433658\\
9	6.00221703238397\\
8	7.99489752633762\\
7	7.95153539757154\\
6	6.83217461278679\\
5	5.78867276088681\\
4	7.29052492785402\\
3	4.18073077261002\\
2	5.0517773310569\\
1	6.80171013837021\\
}--cycle;
\addplot [color=white!55!mycolor1, forget plot]
  table[row sep=crcr]{%
1	5.35186719527366\\
2	3.81430664684546\\
3	3.06648210018055\\
4	5.15890305184559\\
5	4.11788551600075\\
6	5.45209252069895\\
7	6.198042243057\\
8	6.04647066566451\\
9	4.4040130782629\\
10	5.13534041074813\\
11	4.01745200958745\\
12	4.39966058634739\\
};
\addplot [color=white!55!mycolor1, forget plot]
  table[row sep=crcr]{%
1	6.80171013837021\\
2	5.0517773310569\\
3	4.18073077261002\\
4	7.29052492785402\\
5	5.78867276088681\\
6	6.83217461278679\\
7	7.95153539757154\\
8	7.99489752633762\\
9	6.00221703238397\\
10	6.3974943433658\\
11	5.52786786427714\\
12	6.44962198499411\\
};
\addplot [color=mycolor1, mark=triangle, mark options={solid, mycolor1}]
  table[row sep=crcr]{%
1	6.07678866682194\\
2	4.43304198895118\\
3	3.62360643639529\\
4	6.2247139898498\\
5	4.95327913844378\\
6	6.14213356674287\\
7	7.07478882031427\\
8	7.02068409600107\\
9	5.20311505532344\\
10	5.76641737705696\\
11	4.7726599369323\\
12	5.42464128567075\\
};
\addlegendentry{intra-session manual}

\addplot [color=mycolor1, line width=1.5pt, mark=triangle, mark options={solid, mycolor1}, forget plot]
  table[row sep=crcr]{%
1	6.07678866682194\\
2	4.43304198895118\\
3	3.62360643639529\\
4	6.2247139898498\\
5	4.95327913844378\\
6	6.14213356674287\\
7	7.07478882031427\\
8	7.02068409600107\\
9	5.20311505532344\\
10	5.76641737705696\\
11	4.7726599369323\\
12	5.42464128567075\\
};

\addplot[area legend, draw=none, fill=blue, fill opacity=0.2, forget plot]
table[row sep=crcr] {%
x	y\\
1	2.98088081747461\\
2	3.11000389494874\\
3	2.58582496209636\\
4	2.8575183792158\\
5	3.29538137658591\\
6	3.00167032446618\\
7	3.26484745916995\\
8	3.37472831913217\\
9	3.15094343119128\\
10	3.42719903357586\\
11	3.54199520223946\\
12	4.58231554406249\\
12	7.30428487458393\\
11	4.73447338017438\\
10	4.46401140901672\\
9	4.37784722003074\\
8	4.69113684409284\\
7	4.33065622553427\\
6	4.09647448741378\\
5	4.39905415691335\\
4	4.12164611071109\\
3	3.71735835717933\\
2	4.6265216868795\\
1	4.27929037018106\\
}--cycle;
\addplot [color=white!55!blue, forget plot]
  table[row sep=crcr]{%
1	2.98088081747461\\
2	3.11000389494874\\
3	2.58582496209636\\
4	2.8575183792158\\
5	3.29538137658591\\
6	3.00167032446618\\
7	3.26484745916995\\
8	3.37472831913217\\
9	3.15094343119128\\
10	3.42719903357586\\
11	3.54199520223946\\
12	4.58231554406249\\
};
\addplot [color=white!55!blue, forget plot]
  table[row sep=crcr]{%
1	4.27929037018106\\
2	4.6265216868795\\
3	3.71735835717933\\
4	4.12164611071109\\
5	4.39905415691335\\
6	4.09647448741378\\
7	4.33065622553427\\
8	4.69113684409284\\
9	4.37784722003074\\
10	4.46401140901672\\
11	4.73447338017438\\
12	7.30428487458393\\
};
\addplot [color=blue, mark=asterisk, mark options={solid, blue}]
  table[row sep=crcr]{%
1	3.63008559382783\\
2	3.86826279091412\\
3	3.15159165963784\\
4	3.48958224496345\\
5	3.84721776674963\\
6	3.54907240593998\\
7	3.79775184235211\\
8	4.0329325816125\\
9	3.76439532561101\\
10	3.94560522129629\\
11	4.13823429120692\\
12	5.94330020932321\\
};
\addlegendentry{inter-session automatic}

\addplot [color=blue, line width=1.5pt, mark=asterisk, mark options={solid, blue}, forget plot]
  table[row sep=crcr]{%
1	3.63008559382783\\
2	3.86826279091412\\
3	3.15159165963784\\
4	3.48958224496345\\
5	3.84721776674963\\
6	3.54907240593998\\
7	3.79775184235211\\
8	4.0329325816125\\
9	3.76439532561101\\
10	3.94560522129629\\
11	4.13823429120692\\
12	5.94330020932321\\
};

\addplot[area legend, draw=none, fill=green, fill opacity=0.2, forget plot]
table[row sep=crcr] {%
x	y\\
1	1.79095266279206\\
2	2.26905483700378\\
3	2.60254507944634\\
4	2.15454685638868\\
5	2.9998280838513\\
6	2.3057246928314\\
7	2.41213473942856\\
8	2.60612972479828\\
9	2.85309173206092\\
10	2.30588392029248\\
11	2.98215140334215\\
12	2.12923338033663\\
12	3.18980038917808\\
11	4.01522767871278\\
10	2.99330250536204\\
9	4.03233987228529\\
8	4.10593698831703\\
7	3.73683684098704\\
6	3.21306425775682\\
5	4.18134030483733\\
4	3.13117985642939\\
3	3.64225578483685\\
2	2.98421269781436\\
1	2.54793873171575\\
}--cycle;
\addplot [color=white!55!green, forget plot]
  table[row sep=crcr]{%
1	1.79095266279206\\
2	2.26905483700378\\
3	2.60254507944634\\
4	2.15454685638868\\
5	2.9998280838513\\
6	2.3057246928314\\
7	2.41213473942856\\
8	2.60612972479828\\
9	2.85309173206092\\
10	2.30588392029248\\
11	2.98215140334215\\
12	2.12923338033663\\
};
\addplot [color=white!55!green, forget plot]
  table[row sep=crcr]{%
1	2.54793873171575\\
2	2.98421269781436\\
3	3.64225578483685\\
4	3.13117985642939\\
5	4.18134030483733\\
6	3.21306425775682\\
7	3.73683684098704\\
8	4.10593698831703\\
9	4.03233987228529\\
10	2.99330250536204\\
11	4.01522767871278\\
12	3.18980038917808\\
};
\addplot [color=green, mark=triangle, mark options={solid, rotate=180, green}]
  table[row sep=crcr]{%
1	2.16944569725391\\
2	2.62663376740907\\
3	3.1224004321416\\
4	2.64286335640903\\
5	3.59058419434432\\
6	2.75939447529411\\
7	3.0744857902078\\
8	3.35603335655765\\
9	3.44271580217311\\
10	2.64959321282726\\
11	3.49868954102747\\
12	2.65951688475736\\
};
\addlegendentry{intra-session automatic}

\addplot [color=green, line width=1.5pt, mark=triangle, mark options={solid, rotate=180, green}, forget plot]
  table[row sep=crcr]{%
1	2.16944569725391\\
2	2.62663376740907\\
3	3.1224004321416\\
4	2.64286335640903\\
5	3.59058419434432\\
6	2.75939447529411\\
7	3.0744857902078\\
8	3.35603335655765\\
9	3.44271580217311\\
10	2.64959321282726\\
11	3.49868954102747\\
12	2.65951688475736\\
};
\end{axis}
\end{tikzpicture}%
      }}
      \end{tabular}
    };
    
    \def\y{0.5}

    \node[] at (0.8,\y) {\tiny GM};

    \def\y{0.1}

    \node[] at (0.8,\y) {\tiny Accuracy};
    \node[] at (3.9,\y) {\tiny Precision};

    \def\y{-0.5}
    \node[] at (4.5,\y) {\tiny intra};
    \node[] at (5.7,\y) {\tiny inter};
    \node[] at (7.25,\y) {\tiny intra};
    \node[] at (8.5,\y) {\tiny inter};
    \node[] at (10,\y) {\tiny intra};
    \node[] at (11.25,\y) {\tiny inter};
    
    \def\y{-3.2}

    \node[] at (0.8,\y) {\tiny Accuracy};
    \node[] at (6.75,\y) {\tiny Precision};

  \end{tikzpicture}
\end{minipage}}\\
\fbox{\begin{minipage}{\textwidth}
  \begin{tikzpicture}
    \node[anchor=north west,inner sep=0] at (0,0) {
      \begin{tabular}{c}
      \resizebox{\textwidth}{!}{
	\fbox{\resizebox{0.25\textwidth}{!}{
	  \setlength\fheight{5cm} 
	  \setlength\fwidth{1.79cm}
%
%
\begin{tikzpicture}

\begin{axis}[%
width=0.951\fwidth,
height=\fheight,
at={(0\fwidth,0\fheight)},
scale only axis,
xmin=0.5,
xmax=1.5,
xtick={\empty},
ymin=0.85,
ymax=1,
ylabel style={font=\color{white!15!black}},
ylabel={DSC},
axis background/.style={fill=white}
]
\addplot [color=black, dashed, forget plot]
  table[row sep=crcr]{%
1	0.966712240833114\\
1	0.983921161825726\\
};
\addplot [color=black, dashed, forget plot]
  table[row sep=crcr]{%
1	0.909985712017781\\
1	0.94387650405255\\
};
\addplot [color=black, forget plot]
  table[row sep=crcr]{%
0.9625	0.983921161825726\\
1.0375	0.983921161825726\\
};
\addplot [color=black, forget plot]
  table[row sep=crcr]{%
0.9625	0.909985712017781\\
1.0375	0.909985712017781\\
};
\addplot [color=blue, forget plot]
  table[row sep=crcr]{%
0.925	0.94387650405255\\
0.925	0.966712240833114\\
1.075	0.966712240833114\\
1.075	0.94387650405255\\
0.925	0.94387650405255\\
};
\addplot [color=red, forget plot]
  table[row sep=crcr]{%
0.925	0.955818022747157\\
1.075	0.955818022747157\\
};
\addplot [color=black, draw=none, mark=+, mark options={solid, red}, forget plot]
  table[row sep=crcr]{%
1	0.808929358202379\\
1	0.834424174906002\\
1	0.844507845934379\\
1	0.860397006182883\\
1	0.862813197717688\\
1	0.869922275993547\\
1	0.874600198522113\\
1	0.880209519471646\\
1	0.884467406837419\\
1	0.88712027799574\\
1	0.892147587511826\\
1	0.894663702056698\\
1	0.894695989650712\\
1	0.895240700218818\\
1	0.896270965640775\\
1	0.896909648502005\\
1	0.898721484081223\\
1	0.898911353032659\\
1	0.899261083743842\\
1	0.900276158330776\\
1	0.900320718409237\\
1	0.900694507080364\\
1	0.900754147812971\\
1	0.902057107767885\\
1	0.902861643475978\\
1	0.903095718182896\\
1	0.903798170224563\\
1	0.90434280767566\\
1	0.904971871674015\\
1	0.905050967815829\\
1	0.906832298136646\\
1	0.907126922625338\\
1	0.907372400756144\\
1	0.907873195415985\\
1	0.908538350217077\\
1	0.909537856440511\\
};
\end{axis}
\end{tikzpicture}%
%
%
\begin{tikzpicture}

\begin{axis}[%
width=0.951\fwidth,
height=\fheight,
at={(0\fwidth,0\fheight)},
scale only axis,
xmin=0.5,
xmax=1.5,
xtick={\empty},
ymin=0,
ymax=1.4,
ylabel style={font=\color{white!15!black}},
ylabel={HD (mm)},
axis background/.style={fill=white}
]
\addplot [color=black, dashed, forget plot]
  table[row sep=crcr]{%
1	0.480740195134419\\
1	0.754247272602709\\
};
\addplot [color=black, dashed, forget plot]
  table[row sep=crcr]{%
1	0.133333340287209\\
1	0.29814241254931\\
};
\addplot [color=black, forget plot]
  table[row sep=crcr]{%
0.9625	0.754247272602709\\
1.0375	0.754247272602709\\
};
\addplot [color=black, forget plot]
  table[row sep=crcr]{%
0.9625	0.133333340287209\\
1.0375	0.133333340287209\\
};
\addplot [color=blue, forget plot]
  table[row sep=crcr]{%
0.925	0.29814241254931\\
0.925	0.480740195134419\\
1.075	0.480740195134419\\
1.075	0.29814241254931\\
0.925	0.29814241254931\\
};
\addplot [color=red, forget plot]
  table[row sep=crcr]{%
0.925	0.377123636301355\\
1.075	0.377123636301355\\
};
\addplot [color=black, draw=none, mark=+, mark options={solid, red}, forget plot]
  table[row sep=crcr]{%
1	0.777460293193752\\
1	0.777460293193752\\
1	0.777460293193752\\
1	0.777460293193752\\
1	0.777460293193752\\
1	0.777460380081706\\
1	0.777460380081706\\
1	0.800000041723251\\
1	0.800000041723251\\
1	0.800000041723251\\
1	0.800000041723251\\
1	0.800000131130219\\
1	0.811035046338534\\
1	0.811035046338534\\
1	0.811035136978759\\
1	0.853749942850907\\
1	0.853749942850907\\
1	0.894427237647929\\
1	0.894427237647929\\
1	0.894427237647929\\
1	0.894427237647929\\
1	0.894427237647929\\
1	0.894427237647929\\
1	0.894427237647929\\
1	0.894427337607957\\
1	0.942809090753387\\
1	0.942809090753387\\
1	0.942809090753387\\
1	0.942809090753387\\
1	0.942809090753387\\
1	0.942809090753387\\
1	1.01543646707432\\
1	1.06666672229767\\
1	1.07496775583707\\
1	1.09949489084117\\
1	1.09949489084117\\
1	1.09949501371929\\
1	1.19256965019724\\
1	1.25786421654361\\
1	1.33998348603503\\
1	1.35973860787397\\
1	1.46666674315929\\
1	1.52023397941856\\
1	1.52023397941856\\
1	1.56062673615564\\
1	1.6000000834465\\
1	1.60554602757472\\
1	1.68654817338347\\
1	1.68654817338347\\
1	1.87142261061805\\
1	2.26666678488255\\
};
\end{axis}
\end{tikzpicture}%
	}}\,\fbox{\resizebox{0.75\textwidth}{!}{
	  \setlength\fheight{5cm} 
	  \setlength\fwidth{5cm}
%
%
\begin{tikzpicture}

\begin{axis}[%
width=0.951\fwidth,
height=\fheight,
at={(0\fwidth,0\fheight)},
scale only axis,
xmin=0.5,
xmax=4.5,
xtick={1,2,3,4},
xticklabels={{auto},{manual},{auto},{manual}},
ymin=0.85,
ymax=1,
ylabel style={font=\color{white!15!black}},
ylabel={DSC},
axis background/.style={fill=white}
]
\addplot [color=black, dashed, forget plot]
  table[row sep=crcr]{%
1	0.956206978953555\\
1	0.968659315147998\\
};
\addplot [color=black, dashed, forget plot]
  table[row sep=crcr]{%
2	0.945997512837566\\
2	0.966468883676004\\
};
\addplot [color=black, dashed, forget plot]
  table[row sep=crcr]{%
3	0.95048104131296\\
3	0.967642526964561\\
};
\addplot [color=black, dashed, forget plot]
  table[row sep=crcr]{%
4	0.940743767879036\\
4	0.964457509353287\\
};
\addplot [color=black, dashed, forget plot]
  table[row sep=crcr]{%
1	0.909483340078241\\
1	0.936984400128108\\
};
\addplot [color=black, dashed, forget plot]
  table[row sep=crcr]{%
2	0.889836740697161\\
2	0.923387263339071\\
};
\addplot [color=black, dashed, forget plot]
  table[row sep=crcr]{%
3	0.904016692749087\\
3	0.931893483863051\\
};
\addplot [color=black, dashed, forget plot]
  table[row sep=crcr]{%
4	0.873373807458803\\
4	0.913669998701805\\
};
\addplot [color=black, forget plot]
  table[row sep=crcr]{%
0.875	0.968659315147998\\
1.125	0.968659315147998\\
};
\addplot [color=black, forget plot]
  table[row sep=crcr]{%
1.875	0.966468883676004\\
2.125	0.966468883676004\\
};
\addplot [color=black, forget plot]
  table[row sep=crcr]{%
2.875	0.967642526964561\\
3.125	0.967642526964561\\
};
\addplot [color=black, forget plot]
  table[row sep=crcr]{%
3.875	0.964457509353287\\
4.125	0.964457509353287\\
};
\addplot [color=black, forget plot]
  table[row sep=crcr]{%
0.875	0.909483340078241\\
1.125	0.909483340078241\\
};
\addplot [color=black, forget plot]
  table[row sep=crcr]{%
1.875	0.889836740697161\\
2.125	0.889836740697161\\
};
\addplot [color=black, forget plot]
  table[row sep=crcr]{%
2.875	0.904016692749087\\
3.125	0.904016692749087\\
};
\addplot [color=black, forget plot]
  table[row sep=crcr]{%
3.875	0.873373807458803\\
4.125	0.873373807458803\\
};
\addplot [color=blue, forget plot]
  table[row sep=crcr]{%
0.75	0.936984400128108\\
0.75	0.956206978953555\\
1.25	0.956206978953555\\
1.25	0.936984400128108\\
0.75	0.936984400128108\\
};
\addplot [color=blue, forget plot]
  table[row sep=crcr]{%
1.75	0.923387263339071\\
1.75	0.945997512837566\\
2.25	0.945997512837566\\
2.25	0.923387263339071\\
1.75	0.923387263339071\\
};
\addplot [color=blue, forget plot]
  table[row sep=crcr]{%
2.75	0.931893483863051\\
2.75	0.95048104131296\\
3.25	0.95048104131296\\
3.25	0.931893483863051\\
2.75	0.931893483863051\\
};
\addplot [color=blue, forget plot]
  table[row sep=crcr]{%
3.75	0.913669998701805\\
3.75	0.940743767879036\\
4.25	0.940743767879036\\
4.25	0.913669998701805\\
3.75	0.913669998701805\\
};
\addplot [color=red, forget plot]
  table[row sep=crcr]{%
0.75	0.947543252595156\\
1.25	0.947543252595156\\
};
\addplot [color=red, forget plot]
  table[row sep=crcr]{%
1.75	0.936190326138333\\
2.25	0.936190326138333\\
};
\addplot [color=red, forget plot]
  table[row sep=crcr]{%
2.75	0.941102803028514\\
3.25	0.941102803028514\\
};
\addplot [color=red, forget plot]
  table[row sep=crcr]{%
3.75	0.929622485577839\\
4.25	0.929622485577839\\
};
\addplot [color=black, draw=none, mark=+, mark options={solid, red}, forget plot]
  table[row sep=crcr]{%
1	0.820841808793616\\
1	0.843864395989177\\
1	0.878099411832545\\
1	0.888962765957447\\
1	0.895495261586395\\
1	0.89957676029242\\
1	0.900626678603402\\
1	0.902597402597403\\
1	0.903252623921854\\
1	0.90339655932951\\
1	0.906760435571688\\
1	0.90708024275118\\
1	0.907345789608395\\
1	0.907913441794345\\
};
\addplot [color=black, draw=none, mark=+, mark options={solid, red}, forget plot]
  table[row sep=crcr]{%
2	0.865134865134865\\
2	0.868378812199037\\
2	0.870187126481826\\
2	0.870657972732662\\
2	0.874498845826753\\
2	0.877604491578291\\
2	0.88123420667642\\
2	0.883015689165\\
2	0.885815991237678\\
2	0.886675720537196\\
2	0.887997787610619\\
2	0.888818632100327\\
};
\addplot [color=black, draw=none, mark=+, mark options={solid, red}, forget plot]
  table[row sep=crcr]{%
3	0.781627938578611\\
3	0.785852445426913\\
3	0.820464532328939\\
3	0.846363782308557\\
3	0.848978288633461\\
3	0.86395004929346\\
3	0.880859605620498\\
3	0.881830034372407\\
3	0.881845035219268\\
3	0.88212644392001\\
3	0.88715953307393\\
3	0.888136511375948\\
3	0.888543038429307\\
3	0.889133357780223\\
3	0.891040537051381\\
3	0.891554928252176\\
3	0.892656807751147\\
3	0.892967180174146\\
3	0.894303201506591\\
3	0.894776494224008\\
3	0.894960079840319\\
3	0.895912547528517\\
3	0.896021220159151\\
3	0.896999690689762\\
3	0.897263810015488\\
3	0.898574561403509\\
3	0.898636793089486\\
3	0.899583175445244\\
3	0.899836244541485\\
3	0.900153182008077\\
3	0.900549725137431\\
3	0.902022215892908\\
3	0.903246139300347\\
};
\addplot [color=black, draw=none, mark=+, mark options={solid, red}, forget plot]
  table[row sep=crcr]{%
4	0.828263473053892\\
4	0.840208362663663\\
4	0.843819599109131\\
4	0.848839828790268\\
4	0.852244470619029\\
4	0.854078212290503\\
4	0.859435449204757\\
4	0.859901744875487\\
4	0.860103626943005\\
4	0.86084142394822\\
4	0.860913231344155\\
4	0.865433249035006\\
4	0.8666204345816\\
4	0.86723862149193\\
4	0.867884842907848\\
4	0.867962892362435\\
4	0.871357079776273\\
4	0.871753434145215\\
};
\end{axis}
\end{tikzpicture}%
%
%
\begin{tikzpicture}

\begin{axis}[%
width=0.951\fwidth,
height=\fheight,
at={(0\fwidth,0\fheight)},
scale only axis,
xmin=0.5,
xmax=4.5,
xtick={1,2,3,4},
xticklabels={{auto},{manual},{auto},{manual}},
ymin=0,
ymax=1.4,
ylabel style={font=\color{white!15!black}},
ylabel={HD},
axis background/.style={fill=white}
]
\addplot [color=black, dashed, forget plot]
  table[row sep=crcr]{%
1	0.549747445420584\\
1	0.800000041723251\\
};
\addplot [color=black, dashed, forget plot]
  table[row sep=crcr]{%
2	0.596284825098619\\
2	0.853749942850907\\
};
\addplot [color=black, dashed, forget plot]
  table[row sep=crcr]{%
3	0.596284825098619\\
3	0.853749942850907\\
};
\addplot [color=black, dashed, forget plot]
  table[row sep=crcr]{%
4	0.666666701436043\\
4	0.942809090753387\\
};
\addplot [color=black, dashed, forget plot]
  table[row sep=crcr]{%
1	0.266666680574417\\
1	0.377123678448203\\
};
\addplot [color=black, dashed, forget plot]
  table[row sep=crcr]{%
2	0.266666680574417\\
2	0.421637043345868\\
};
\addplot [color=black, dashed, forget plot]
  table[row sep=crcr]{%
3	0.266666680574417\\
3	0.400000020861626\\
};
\addplot [color=black, dashed, forget plot]
  table[row sep=crcr]{%
4	0.26666671037674\\
4	0.480740195134419\\
};
\addplot [color=black, forget plot]
  table[row sep=crcr]{%
0.875	0.800000041723251\\
1.125	0.800000041723251\\
};
\addplot [color=black, forget plot]
  table[row sep=crcr]{%
1.875	0.853749942850907\\
2.125	0.853749942850907\\
};
\addplot [color=black, forget plot]
  table[row sep=crcr]{%
2.875	0.853749942850907\\
3.125	0.853749942850907\\
};
\addplot [color=black, forget plot]
  table[row sep=crcr]{%
3.875	0.942809090753387\\
4.125	0.942809090753387\\
};
\addplot [color=black, forget plot]
  table[row sep=crcr]{%
0.875	0.266666680574417\\
1.125	0.266666680574417\\
};
\addplot [color=black, forget plot]
  table[row sep=crcr]{%
1.875	0.266666680574417\\
2.125	0.266666680574417\\
};
\addplot [color=black, forget plot]
  table[row sep=crcr]{%
2.875	0.266666680574417\\
3.125	0.266666680574417\\
};
\addplot [color=black, forget plot]
  table[row sep=crcr]{%
3.875	0.26666671037674\\
4.125	0.26666671037674\\
};
\addplot [color=blue, forget plot]
  table[row sep=crcr]{%
0.75	0.377123678448203\\
0.75	0.549747445420584\\
1.25	0.549747445420584\\
1.25	0.377123678448203\\
0.75	0.377123678448203\\
};
\addplot [color=blue, forget plot]
  table[row sep=crcr]{%
1.75	0.421637043345868\\
1.75	0.596284825098619\\
2.25	0.596284825098619\\
2.25	0.421637043345868\\
1.75	0.421637043345868\\
};
\addplot [color=blue, forget plot]
  table[row sep=crcr]{%
2.75	0.400000020861626\\
2.75	0.596284825098619\\
3.25	0.596284825098619\\
3.25	0.400000020861626\\
2.75	0.400000020861626\\
};
\addplot [color=blue, forget plot]
  table[row sep=crcr]{%
3.75	0.480740195134419\\
3.75	0.666666701436043\\
4.25	0.666666701436043\\
4.25	0.480740195134419\\
3.75	0.480740195134419\\
};
\addplot [color=red, forget plot]
  table[row sep=crcr]{%
0.75	0.480740195134419\\
1.25	0.480740195134419\\
};
\addplot [color=red, forget plot]
  table[row sep=crcr]{%
1.75	0.533333361148834\\
2.25	0.533333361148834\\
};
\addplot [color=red, forget plot]
  table[row sep=crcr]{%
2.75	0.480740195134419\\
3.25	0.480740195134419\\
};
\addplot [color=red, forget plot]
  table[row sep=crcr]{%
3.75	0.549747445420584\\
4.25	0.549747445420584\\
};
\addplot [color=black, draw=none, mark=+, mark options={solid, red}, forget plot]
  table[row sep=crcr]{%
1	0.811035046338534\\
1	0.942809090753387\\
1	0.961480390268838\\
1	1.01543658055819\\
1	1.04136667776572\\
1	1.07496775583707\\
1	1.31318110872721\\
1	1.33333340287209\\
1	1.33333340287209\\
1	1.37275075705946\\
1	1.46666674315929\\
1	1.6000000834465\\
1	1.77888863465978\\
1	1.90904290800119\\
};
\addplot [color=black, draw=none, mark=+, mark options={solid, red}, forget plot]
  table[row sep=crcr]{%
2	0.894427337607957\\
2	0.93333338201046\\
2	0.942809090753387\\
2	0.970681369195712\\
2	1.01543646707432\\
};
\addplot [color=black, draw=none, mark=+, mark options={solid, red}, forget plot]
  table[row sep=crcr]{%
3	0.894427237647929\\
3	0.894427237647929\\
3	0.93333338201046\\
3	0.93333338201046\\
3	0.93333338201046\\
3	0.942809090753387\\
3	0.942809090753387\\
3	0.942809090753387\\
3	0.961480390268838\\
3	0.961480390268838\\
3	0.970681369195712\\
3	1.04136667776572\\
3	1.04136667776572\\
3	1.06666672229767\\
3	1.06666684150696\\
3	1.13920055878961\\
3	1.20000006258488\\
3	1.31318110872721\\
3	1.33998348603503\\
3	1.52023397941856\\
3	1.5549205863875\\
3	1.56062673615564\\
3	1.73333342373371\\
3	1.77888863465978\\
3	1.79381663316465\\
3	1.90904290800119\\
3	2.03960791181095\\
3	2.13333344459534\\
};
\addplot [color=black, draw=none, mark=+, mark options={solid, red}, forget plot]
  table[row sep=crcr]{%
4	0.961480390268838\\
4	0.961480390268838\\
4	0.970681369195712\\
4	1.01543646707432\\
4	1.01543646707432\\
4	1.04136667776572\\
4	1.06666672229767\\
4	1.07496775583707\\
4	1.09949489084117\\
4	1.13920055878961\\
4	1.33333340287209\\
4	1.33998348603503\\
};
\end{axis}
\end{tikzpicture}%
%
%
\begin{tikzpicture}

\begin{axis}[%
width=0.951\fwidth,
height=\fheight,
at={(0\fwidth,0\fheight)},
scale only axis,
xmin=0.5,
xmax=4.5,
xtick={1,2,3,4},
xticklabels={{auto},{manual},{auto},{manual}},
ymin=0,
ymax=12,
ylabel style={font=\color{white!15!black}},
ylabel={RSD},
axis background/.style={fill=white}
]
\addplot [color=black, dashed, forget plot]
  table[row sep=crcr]{%
1	2.80141961023598\\
1	5.85502430281854\\
};
\addplot [color=black, dashed, forget plot]
  table[row sep=crcr]{%
2	5.69174838316246\\
2	11.8180981421822\\
};
\addplot [color=black, dashed, forget plot]
  table[row sep=crcr]{%
3	3.76337270756144\\
3	8.00498242852696\\
};
\addplot [color=black, dashed, forget plot]
  table[row sep=crcr]{%
4	6.64279057712079\\
4	13.8564192319401\\
};
\addplot [color=black, dashed, forget plot]
  table[row sep=crcr]{%
1	0.0184840355819192\\
1	0.736733019272106\\
};
\addplot [color=black, dashed, forget plot]
  table[row sep=crcr]{%
2	0.0178584867075864\\
2	1.28213029580332\\
};
\addplot [color=black, dashed, forget plot]
  table[row sep=crcr]{%
3	0.0145121966379985\\
3	0.883413034598802\\
};
\addplot [color=black, dashed, forget plot]
  table[row sep=crcr]{%
4	0.0370212974443288\\
4	1.72965974685127\\
};
\addplot [color=black, forget plot]
  table[row sep=crcr]{%
0.875	5.85502430281854\\
1.125	5.85502430281854\\
};
\addplot [color=black, forget plot]
  table[row sep=crcr]{%
1.875	11.8180981421822\\
2.125	11.8180981421822\\
};
\addplot [color=black, forget plot]
  table[row sep=crcr]{%
2.875	8.00498242852696\\
3.125	8.00498242852696\\
};
\addplot [color=black, forget plot]
  table[row sep=crcr]{%
3.875	13.8564192319401\\
4.125	13.8564192319401\\
};
\addplot [color=black, forget plot]
  table[row sep=crcr]{%
0.875	0.0184840355819192\\
1.125	0.0184840355819192\\
};
\addplot [color=black, forget plot]
  table[row sep=crcr]{%
1.875	0.0178584867075864\\
2.125	0.0178584867075864\\
};
\addplot [color=black, forget plot]
  table[row sep=crcr]{%
2.875	0.0145121966379985\\
3.125	0.0145121966379985\\
};
\addplot [color=black, forget plot]
  table[row sep=crcr]{%
3.875	0.0370212974443288\\
4.125	0.0370212974443288\\
};
\addplot [color=blue, forget plot]
  table[row sep=crcr]{%
0.75	0.736733019272106\\
0.75	2.80141961023598\\
1.25	2.80141961023598\\
1.25	0.736733019272106\\
0.75	0.736733019272106\\
};
\addplot [color=blue, forget plot]
  table[row sep=crcr]{%
1.75	1.28213029580332\\
1.75	5.69174838316246\\
2.25	5.69174838316246\\
2.25	1.28213029580332\\
1.75	1.28213029580332\\
};
\addplot [color=blue, forget plot]
  table[row sep=crcr]{%
2.75	0.883413034598802\\
2.75	3.76337270756144\\
3.25	3.76337270756144\\
3.25	0.883413034598802\\
2.75	0.883413034598802\\
};
\addplot [color=blue, forget plot]
  table[row sep=crcr]{%
3.75	1.72965974685127\\
3.75	6.64279057712079\\
4.25	6.64279057712079\\
4.25	1.72965974685127\\
3.75	1.72965974685127\\
};
\addplot [color=red, forget plot]
  table[row sep=crcr]{%
0.75	1.48939241774498\\
1.25	1.48939241774498\\
};
\addplot [color=red, forget plot]
  table[row sep=crcr]{%
1.75	2.80006285170298\\
2.25	2.80006285170298\\
};
\addplot [color=red, forget plot]
  table[row sep=crcr]{%
2.75	1.94960238966367\\
3.25	1.94960238966367\\
};
\addplot [color=red, forget plot]
  table[row sep=crcr]{%
3.75	3.64964918028074\\
4.25	3.64964918028074\\
};
\addplot [color=black, draw=none, mark=+, mark options={solid, red}, forget plot]
  table[row sep=crcr]{%
1	6.10240320649686\\
1	6.37209770929233\\
1	6.43969035015986\\
1	6.66023279950472\\
1	6.92141558304292\\
1	7.01464088988064\\
1	7.05839889077947\\
1	7.7598135780547\\
1	7.89893719048345\\
1	8.44904393973407\\
1	9.37302538799841\\
1	9.48150735868648\\
1	9.67231831476149\\
1	9.70177123523895\\
1	10.9780477095781\\
1	11.8830836485372\\
1	11.9053692750796\\
1	13.7284029326112\\
1	14.0426464906382\\
};
\addplot [color=black, draw=none, mark=+, mark options={solid, red}, forget plot]
  table[row sep=crcr]{%
2	12.3854270878073\\
2	12.8979594691505\\
2	13.2382421496229\\
2	13.4535257084773\\
2	13.7967864243178\\
2	14.2361787177663\\
2	14.463761530322\\
};
\addplot [color=black, draw=none, mark=+, mark options={solid, red}, forget plot]
  table[row sep=crcr]{%
3	8.10634568822114\\
3	8.18014301614784\\
3	8.32624999925284\\
3	8.3618617553173\\
3	8.94135479508121\\
3	9.02193680327469\\
3	9.07294163029237\\
3	9.11537323195079\\
3	9.14459081402479\\
3	9.19842573216052\\
3	9.26374402153695\\
3	9.28663514907417\\
3	9.41256453617333\\
3	10.0210840902399\\
3	10.0398496646384\\
3	10.2283459069015\\
3	10.2773740509567\\
3	10.3757045763772\\
3	10.7544757594912\\
3	11.3915041369935\\
3	11.3944794928582\\
3	11.5757885679278\\
3	11.7288786699882\\
3	11.8393838200794\\
3	11.8642077380293\\
3	12.176271881096\\
3	12.4337026070851\\
3	12.9659171452981\\
3	15.1433688570052\\
};
\addplot [color=black, draw=none, mark=+, mark options={solid, red}, forget plot]
  table[row sep=crcr]{%
4	14.1666666482301\\
4	14.1711959773637\\
4	14.5633585760606\\
4	14.6109959412486\\
4	14.785801241039\\
4	14.8392026803318\\
4	14.9890795494951\\
4	15.391084264899\\
4	15.4955438873058\\
4	15.5018394499988\\
4	16.7448232247379\\
4	17.1081929036799\\
4	17.7000683788495\\
4	18.3863511589152\\
4	18.8983175485835\\
};
\end{axis}
\end{tikzpicture}%
	}}
      }\\[4mm]
      \fbox{\resizebox{0.47\textwidth}{!}{
	\setlength\fheight{6.1cm} 
	\setlength\fwidth{5cm}
	\input{figs/MDGRU_plots/WM_55000_accDSC_slice.tex}
	\input{figs/MDGRU_plots/WM_55000_accHD_slice.tex}
      }}\,\fbox{\resizebox{0.47\textwidth}{!}{
	\setlength\fheight{5.765cm} 
	\setlength\fwidth{10cm}
%
%
\definecolor{mycolor1}{rgb}{0.00000,1.00000,1.00000}%
\begin{tikzpicture}

\begin{axis}[%
width=0.951\fwidth,
height=\fheight,
at={(0\fwidth,0\fheight)},
scale only axis,
xmin=0,
xmax=13,
xtick={0,1,2,3,4,5,6,7,8,9,10,11,12,13},
xticklabels={{rostral},{1},{2},{3},{4},{5},{6},{7},{8},{9},{10},{11},{12},{caudal}},
xlabel style={font=\color{white!15!black}},
xlabel={Slice},
ymin=-1,
ymax=12,
ylabel style={font=\color{white!15!black}},
ylabel={RSD},
axis background/.style={fill=white},
axis x line*=bottom,
axis y line*=left,
legend style={at={(0.03,0.97)}, anchor=north west, legend cell align=left, align=left, draw=white!15!black}
]

\addplot[area legend, draw=none, fill=red, fill opacity=0.2, forget plot]
table[row sep=crcr] {%
x	y\\
1	4.0968974870098\\
2	3.61489713217836\\
3	3.88019999947165\\
4	4.5380373943693\\
5	3.02704054631444\\
6	3.77563046180726\\
7	4.89603181357321\\
8	5.22706214413872\\
9	2.83442370433146\\
10	3.82101375139795\\
11	3.30384723931623\\
12	3.3637526050589\\
12	4.95769800497273\\
11	4.57670951830134\\
10	5.28115521522522\\
9	4.18028353273982\\
8	7.05737900028458\\
7	6.40942145236802\\
6	5.21220639517931\\
5	4.24599614405292\\
4	6.25880315804718\\
3	5.32950119519768\\
2	5.02971839177042\\
1	5.57896831981684\\
}--cycle;
\addplot [color=white!55!red, forget plot]
  table[row sep=crcr]{%
1	4.0968974870098\\
2	3.61489713217836\\
3	3.88019999947165\\
4	4.5380373943693\\
5	3.02704054631444\\
6	3.77563046180726\\
7	4.89603181357321\\
8	5.22706214413872\\
9	2.83442370433146\\
10	3.82101375139795\\
11	3.30384723931623\\
12	3.3637526050589\\
};
\addplot [color=white!55!red, forget plot]
  table[row sep=crcr]{%
1	5.57896831981684\\
2	5.02971839177042\\
3	5.32950119519768\\
4	6.25880315804718\\
5	4.24599614405292\\
6	5.21220639517931\\
7	6.40942145236802\\
8	7.05737900028458\\
9	4.18028353273982\\
10	5.28115521522522\\
11	4.57670951830134\\
12	4.95769800497273\\
};
\addplot [color=red, mark=o, mark options={solid, red}]
  table[row sep=crcr]{%
1	4.83793290341332\\
2	4.32230776197439\\
3	4.60485059733467\\
4	5.39842027620824\\
5	3.63651834518368\\
6	4.49391842849329\\
7	5.65272663297061\\
8	6.14222057221165\\
9	3.50735361853564\\
10	4.55108448331158\\
11	3.94027837880878\\
12	4.16072530501581\\
};
\addlegendentry{inter-session manual}

\addplot [color=red, line width=1.5pt, mark=o, mark options={solid, red}, forget plot]
  table[row sep=crcr]{%
1	4.83793290341332\\
2	4.32230776197439\\
3	4.60485059733467\\
4	5.39842027620824\\
5	3.63651834518368\\
6	4.49391842849329\\
7	5.65272663297061\\
8	6.14222057221165\\
9	3.50735361853564\\
10	4.55108448331158\\
11	3.94027837880878\\
12	4.16072530501581\\
};

\addplot[area legend, draw=none, fill=mycolor1, fill opacity=0.2, forget plot]
table[row sep=crcr] {%
x	y\\
1	1.93230688596885\\
2	2.02166422906106\\
3	4.19932678467193\\
4	3.16076426786067\\
5	1.91641902433152\\
6	2.16197305167621\\
7	5.17003193190891\\
8	5.31307639386996\\
9	2.6196098083648\\
10	2.87878574214079\\
11	2.2693307796852\\
12	4.59570674147788\\
12	5.8482751176294\\
11	3.20942802185621\\
10	4.44478123097221\\
9	3.8235942950621\\
8	6.93193050175129\\
7	6.81655213894008\\
6	2.96163240995284\\
5	2.80229808425959\\
4	4.57272670636285\\
3	5.5468594760349\\
2	2.96460993671614\\
1	2.54362545552326\\
}--cycle;
\addplot [color=white!55!mycolor1, forget plot]
  table[row sep=crcr]{%
1	1.93230688596885\\
2	2.02166422906106\\
3	4.19932678467193\\
4	3.16076426786067\\
5	1.91641902433152\\
6	2.16197305167621\\
7	5.17003193190891\\
8	5.31307639386996\\
9	2.6196098083648\\
10	2.87878574214079\\
11	2.2693307796852\\
12	4.59570674147788\\
};
\addplot [color=white!55!mycolor1, forget plot]
  table[row sep=crcr]{%
1	2.54362545552326\\
2	2.96460993671614\\
3	5.5468594760349\\
4	4.57272670636285\\
5	2.80229808425959\\
6	2.96163240995284\\
7	6.81655213894008\\
8	6.93193050175129\\
9	3.8235942950621\\
10	4.44478123097221\\
11	3.20942802185621\\
12	5.8482751176294\\
};
\addplot [color=mycolor1, mark=triangle, mark options={solid, mycolor1}]
  table[row sep=crcr]{%
1	2.23796617074605\\
2	2.4931370828886\\
3	4.87309313035342\\
4	3.86674548711176\\
5	2.35935855429555\\
6	2.56180273081453\\
7	5.99329203542449\\
8	6.12250344781063\\
9	3.22160205171345\\
10	3.6617834865565\\
11	2.7393794007707\\
12	5.22199092955364\\
};
\addlegendentry{intra-session manual}

\addplot [color=mycolor1, line width=1.5pt, mark=triangle, mark options={solid, mycolor1}, forget plot]
  table[row sep=crcr]{%
1	2.23796617074605\\
2	2.4931370828886\\
3	4.87309313035342\\
4	3.86674548711176\\
5	2.35935855429555\\
6	2.56180273081453\\
7	5.99329203542449\\
8	6.12250344781063\\
9	3.22160205171345\\
10	3.6617834865565\\
11	2.7393794007707\\
12	5.22199092955364\\
};

\addplot[area legend, draw=none, fill=blue, fill opacity=0.2, forget plot]
table[row sep=crcr] {%
x	y\\
1	1.78243706006274\\
2	1.75675690748453\\
3	1.29143737781464\\
4	2.04347547544279\\
5	2.18484221846046\\
6	2.31206564997936\\
7	2.94599121453404\\
8	3.06346143782823\\
9	2.1290194291306\\
10	1.88309831891909\\
11	2.45982754492477\\
12	2.79110674356586\\
12	4.06299043105454\\
11	3.3232773820973\\
10	2.70985864802965\\
9	3.21571303684413\\
8	4.40240342653122\\
7	4.21945986823447\\
6	3.19126432866448\\
5	3.34900896552668\\
4	2.82168445850603\\
3	1.91527362520065\\
2	2.56110239841656\\
1	2.54263043007633\\
}--cycle;
\addplot [color=white!55!blue, forget plot]
  table[row sep=crcr]{%
1	1.78243706006274\\
2	1.75675690748453\\
3	1.29143737781464\\
4	2.04347547544279\\
5	2.18484221846046\\
6	2.31206564997936\\
7	2.94599121453404\\
8	3.06346143782823\\
9	2.1290194291306\\
10	1.88309831891909\\
11	2.45982754492477\\
12	2.79110674356586\\
};
\addplot [color=white!55!blue, forget plot]
  table[row sep=crcr]{%
1	2.54263043007633\\
2	2.56110239841656\\
3	1.91527362520065\\
4	2.82168445850603\\
5	3.34900896552668\\
6	3.19126432866448\\
7	4.21945986823447\\
8	4.40240342653122\\
9	3.21571303684413\\
10	2.70985864802965\\
11	3.3232773820973\\
12	4.06299043105454\\
};
\addplot [color=blue, mark=asterisk, mark options={solid, blue}]
  table[row sep=crcr]{%
1	2.16253374506954\\
2	2.15892965295054\\
3	1.60335550150764\\
4	2.43257996697441\\
5	2.76692559199357\\
6	2.75166498932192\\
7	3.58272554138425\\
8	3.73293243217972\\
9	2.67236623298736\\
10	2.29647848347437\\
11	2.89155246351103\\
12	3.4270485873102\\
};
\addlegendentry{inter-session automatic}

\addplot [color=blue, line width=1.5pt, mark=asterisk, mark options={solid, blue}, forget plot]
  table[row sep=crcr]{%
1	2.16253374506954\\
2	2.15892965295054\\
3	1.60335550150764\\
4	2.43257996697441\\
5	2.76692559199357\\
6	2.75166498932192\\
7	3.58272554138425\\
8	3.73293243217972\\
9	2.67236623298736\\
10	2.29647848347437\\
11	2.89155246351103\\
12	3.4270485873102\\
};

\addplot[area legend, draw=none, fill=green, fill opacity=0.2, forget plot]
table[row sep=crcr] {%
x	y\\
1	1.06377895629764\\
2	0.758822436695349\\
3	1.53065899936982\\
4	1.66998297600561\\
5	1.54068200610849\\
6	1.12314867142497\\
7	3.36753010518272\\
8	3.8738565476307\\
9	1.69000366505648\\
10	1.36349812838236\\
11	1.67762341750327\\
12	1.59265219430107\\
12	2.20758456905331\\
11	2.7774190555065\\
10	1.91612470603784\\
9	2.55878972929201\\
8	5.40114660583382\\
7	4.63983111969572\\
6	1.56531252585347\\
5	1.98455817902703\\
4	2.60958318886531\\
3	2.30497306526777\\
2	1.12723355373487\\
1	1.45276045586467\\
}--cycle;
\addplot [color=white!55!green, forget plot]
  table[row sep=crcr]{%
1	1.06377895629764\\
2	0.758822436695349\\
3	1.53065899936982\\
4	1.66998297600561\\
5	1.54068200610849\\
6	1.12314867142497\\
7	3.36753010518272\\
8	3.8738565476307\\
9	1.69000366505648\\
10	1.36349812838236\\
11	1.67762341750327\\
12	1.59265219430107\\
};
\addplot [color=white!55!green, forget plot]
  table[row sep=crcr]{%
1	1.45276045586467\\
2	1.12723355373487\\
3	2.30497306526777\\
4	2.60958318886531\\
5	1.98455817902703\\
6	1.56531252585347\\
7	4.63983111969572\\
8	5.40114660583382\\
9	2.55878972929201\\
10	1.91612470603784\\
11	2.7774190555065\\
12	2.20758456905331\\
};
\addplot [color=green, mark=triangle, mark options={solid, rotate=180, green}]
  table[row sep=crcr]{%
1	1.25826970608116\\
2	0.94302799521511\\
3	1.9178160323188\\
4	2.13978308243546\\
5	1.76262009256776\\
6	1.34423059863922\\
7	4.00368061243922\\
8	4.63750157673226\\
9	2.12439669717424\\
10	1.6398114172101\\
11	2.22752123650489\\
12	1.90011838167719\\
};
\addlegendentry{intra-session automatic}

\addplot [color=green, line width=1.5pt, mark=triangle, mark options={solid, rotate=180, green}, forget plot]
  table[row sep=crcr]{%
1	1.25826970608116\\
2	0.94302799521511\\
3	1.9178160323188\\
4	2.13978308243546\\
5	1.76262009256776\\
6	1.34423059863922\\
7	4.00368061243922\\
8	4.63750157673226\\
9	2.12439669717424\\
10	1.6398114172101\\
11	2.22752123650489\\
12	1.90011838167719\\
};
\end{axis}
\end{tikzpicture}%
      }}
      \end{tabular}
    };
    
    \def\y{0.5}

    \node[] at (0.8,\y) {\tiny WM};

    \def\y{0.1}

    \node[] at (0.8,\y) {\tiny Accuracy};
    \node[] at (3.9,\y) {\tiny Precision};

    \def\y{-0.5}
    \node[] at (4.5,\y) {\tiny intra};
    \node[] at (5.7,\y) {\tiny inter};
    \node[] at (7.25,\y) {\tiny intra};
    \node[] at (8.5,\y) {\tiny inter};
    \node[] at (10,\y) {\tiny intra};
    \node[] at (11.25,\y) {\tiny inter};
    
    \def\y{-3.2}

    \node[] at (0.8,\y) {\tiny Accuracy};
    \node[] at (6.75,\y) {\tiny Precision};

  \end{tikzpicture}
\end{minipage}}
\end{tabular}
}
 
 \caption{GM and WM accuracy and precision plots of the AMIRA dataset.
 For both boxes GM and WM:
 \emph{Top row:} Accuracy (\emph{left}) in DSC and HD of all the 855 slices of the proposed method;
 intra-session (intra) and inter-session (inter) precision (\emph{right}) of the proposed method (auto) and the manual segmentations in DSC, HD, and area RSD.
 \emph{Bottom row:} Accuracy box plots (\emph{left}) in DSC and HD wrt. the slice positions with overlaid error bars in the format mean $\pm$ one standard deviation;
 precision error bars (\emph{right}) for area RSD wrt. the slice positions, for better visualization shown with 0.2 standard deviations.
 HD is measured in millimeters, and RSD in percents.
 }
 \label{fig:accPrecBoxPlots}
\end{figure}

In \figref{fig:accPrecBoxPlots} and in \tabref{tab:CE_GDL} we show GM and WM accuracy and precision of all gathered slice results in DSC, HD and relative standard deviation of the areas (RSD),
also known as coefficient of variation.
With intra- and inter-session precision we compare segmentations of the same slice for different scans with and without repositioning, respectively.
The proposed automatic segmentations shows better reproducibility as the manual segmentations.
Additionally, we show the anatomical GM and WM areas wrt. the slice positions in \figref{fig:areaSlicePlots} and show randomly chosen results in \figref{fig:AMIRAresults}.
Training multiple networks with data from multiple human raters as ground truth data, as we did with the SCGM data, cf. Subsec.~\ref{seq:challengeModel}, might further improve the performance.

\subsection{SCGM challenge model}
\label{seq:challengeModel}
To enable comparison with other methods, we tested MD-GRU on the SCGM dataset \cite{prados_spinal_2017}.
We trained four MD-GRU models, one for each expert rater's ground truth, and in the end performed majority voting on the individual test results to mimic the challenge's consensus segmentation.

We used the same MD-GRU setup but with a window size of $200\times200$ pixels for a similar anatomical field of view as the AMIRA models.
Random subsamples in each training iteration were drawn with a distance of 200 pixels from the image boundary.
We trained the networks for 100'000 iterations and observed, that the scores reached their upper bounds after around 60'000 iterations.
One training iteration took around 4 seconds and segmentation of one slice took less than 1 second.

\begin{figure}

\begin{tikzpicture}
\node[anchor=north west,inner sep=0] at (0,0) {
  \begin{tabular}{c}
  \resizebox{\textwidth}{!}{
    \includegraphics[height=0.15\textwidth]{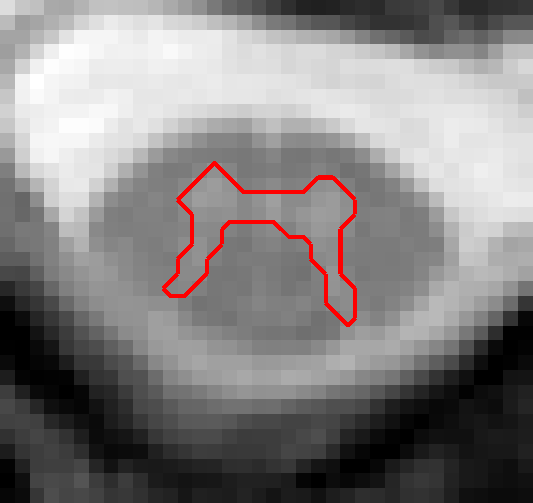}\,
    \includegraphics[height=0.15\textwidth]{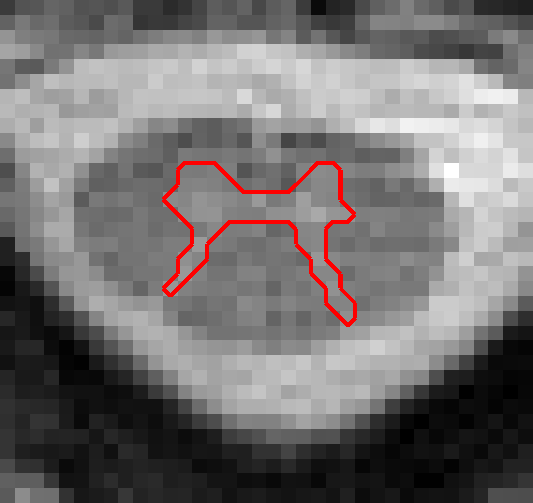}\,
    \includegraphics[height=0.15\textwidth]{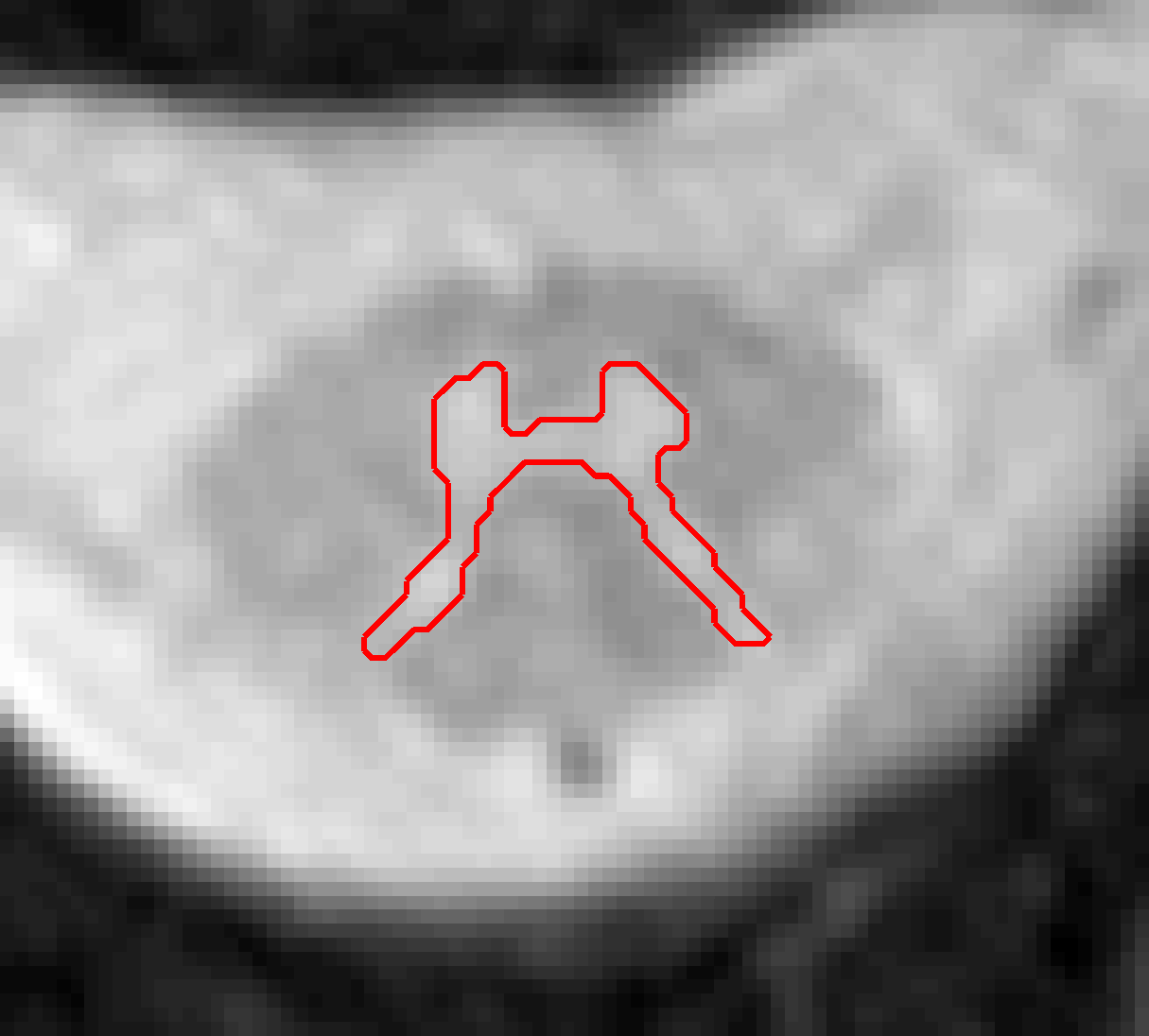}\,
    \includegraphics[height=0.15\textwidth]{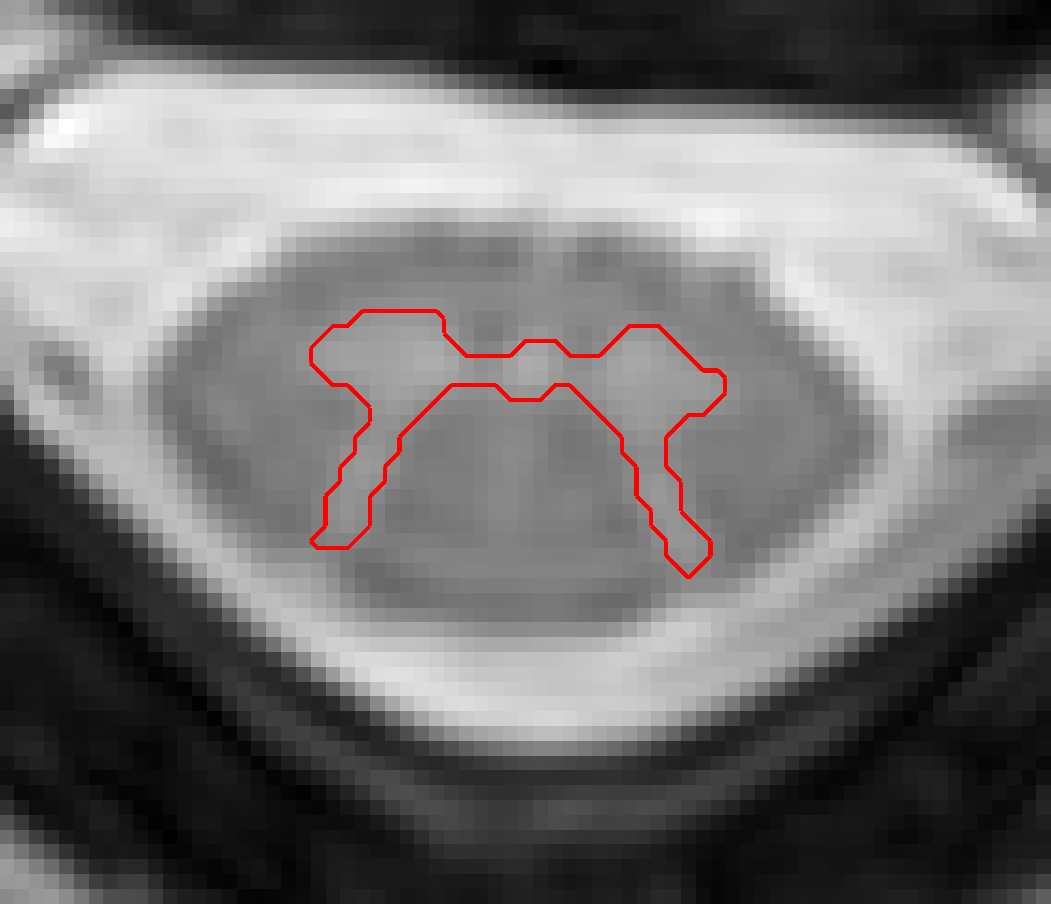}
  }
  \end{tabular}
};

\def\y{-2.8}

\node[] at (1.5,\y) {\tiny Site 1 Subject 15 Slice 2};
\node[] at (4.5,\y) {\tiny Site 2 Subject 13 Slice 5};
\node[] at (7.5,\y) {\tiny Site 3 Subject 12 Slice 14};
\node[] at (10.5,\y) {\tiny Site 4 Subject 19 Slice 7};

\end{tikzpicture}
  
  \caption{
    For each site of the SCGM dataset, one randomly chosen result of the proposed model in cropped view.
  }
  \label{fig:SCGMresults}
\end{figure}

In \tabref{tab:GMchallengeResults}, the proposed model shows a new state-of-the-art in almost all metrics.
This comparison shows MD-GRU's strong performance in learning the GM segmentation problem.
In \tabref{tab:GMchallengeMDGRU}, we additionally show the improvement for the auto-weighted GDL, compared to the native MD-GRU approach with only CEL.
Figure~\ref{fig:SCGMresults} shows randomly chosen results of the proposed model.

\begin{table}
\caption{
Results of the SCGM challenge competitors including the results of Porisky \etal \cite{porisky_grey_2017}, Perone \etal \cite{perone_spinal_2018} and ours.
The metrics are Dice coefficient (DSC), 
mean surface distance (MD), 
Hausdorff surface distance (HD), 
skeletonized Hausdorff distance (SHD), 
skeletonized median distance (SMD), 
true positive rate (TPR), 
true negative rate (TNR), 
precision (P), 
Jaccard index (J), and
conformity (C).
Best results on each metric are highlighted in bold font.
Distances are measured in millimeters.
}
\label{tab:GMchallengeResults}

\resizebox{\textwidth}{!}{%
  \begin{tabular}[] {lrlrlrlrlrlrlrlrlrl}
    \rowcolor{gray}
    & \multicolumn{2}{c}{\bf JCSCS} & \multicolumn{2}{c}{\bf DEEPSEG} & \multicolumn{2}{c}{\bf MGAC} & \multicolumn{2}{c}{\bf GSBME} & \multicolumn{2}{c}{\bf SCT} & \multicolumn{2}{c}{\bf VBEM} & \multicolumn{2}{c}{\bf \cite{porisky_grey_2017}} & \multicolumn{2}{c}{\bf \cite{perone_spinal_2018}} & \multicolumn{2}{c}{\bf Proposed} \tnhl
    DSC & 0.79 & $\pm$ 0.04 & 0.80 & $\pm$ 0.06 & 0.75 & $\pm$ 0.07 & 0.76 & $\pm$ 0.06 & 0.69 & $\pm$ 0.07 & 0.61 & $\pm$ 0.13		& 0.80 & $\pm$ 0.06 & 0.85 & $\pm$ 0.04 & \bf 0.90 & $\pm$ 0.03 \tn
    \rowcolor{gray}
    MD& 0.39 & $\pm$ 0.44 & 0.46 & $\pm$ 0.48 & 0.70 & $\pm$ 0.79 & 0.62 & $\pm$ 0.64 & 0.69 & $\pm$ 0.76 & 1.04 & $\pm$ 1.14		& 0.53 & $\pm$ 0.57 & 0.36 & $\pm$ 0.34 & \bf 0.21 & $\pm$ 0.20 \tn
    HD & 2.65 & $\pm$ 3.40 & 4.07 & $\pm$ 3.27 & 3.56 & $\pm$ 1.34 & 4.92 & $\pm$ 3.30 & 3.26 & $\pm$ 1.35 & 5.34 & $\pm$ 15.35		& 3.69 & $\pm$ 3.93 & 2.61 & $\pm$ 2.15 & \bf 1.85 &  $\pm$ 1.16 \tn
    \rowcolor{gray}
    SHD & 1.00 & $\pm$ 0.35 & 1.26 & $\pm$ 0.65 & 1.07 & $\pm$ 0.37 & 1.86 & $\pm$ 0.85 & 1.12 & $\pm$ 0.41 & 2.77 & $\pm$ 8.10		& 1.22 & $\pm$ 0.51 & 0.85 & $\pm$ 0.32 & \bf 0.71 & $\pm$ 0.28 \tn
    SMD & 0.37 & $\pm$ 0.18 & 0.45 & $\pm$ 0.20 & 0.39 & $\pm$ 0.17 & 0.61 & $\pm$ 0.35 & 0.39 & $\pm$ 0.16 & 0.54 & $\pm$ 0.25		& 0.44 & $\pm$ 0.19 & \bf 0.36 & $\pm$ 0.17 & 0.37 & $\pm$ 0.17 \tn
    \rowcolor{gray}
    TPR & 77.98 & $\pm$ 4.88 & 78.89 & $\pm$ 10.33 & 87.51 & $\pm$ 6.65 & 75.69 & $\pm$ 8.08 & 70.29 & $\pm$ 6.76 & 65.66 & $\pm$ 14.39 	& 79.65 & $\pm$ 9.56 & 94.97 & $\pm$ 3.50 & \bf 96.22 & $\pm$ 2.69 \tn
    TNR & \bf{99.98} & $\pm$ 0.03 & 99.97 & $\pm$ 0.04 & 99.94 & $\pm$ 0.08 & 99.97 & $\pm$ 0.05 & 99.95 & $\pm$ 0.06 & 99.93 & $\pm$ 0.09	& 99.97 & $\pm$ 0.04 & 99.95 & $\pm$ 0.06 & \bf 99.98 & $\pm$ 0.03 \tn
    \rowcolor{gray}
    P & 81.06 & $\pm$ 5.97 & 82.78 & $\pm$ 5.19 & 65.60 & $\pm$ 9.01 & 76.26 & $\pm$ 7.41 & 67.87 & $\pm$ 8.62 & 59.07 & $\pm$ 13.69	& 81.29 & $\pm$ 5.30 & 77.29 & $\pm$ 6.46 & \bf 85.46 & $\pm$ 4.96 \tn
    J & 0.66 & $\pm$ 0.05 & 0.68 & $\pm$ 0.08 & 0.60 & $\pm$ 0.08 & 0.61 & $\pm$ 0.08 & 0.53 & $\pm$ 0.08 & 0.45 & $\pm$ 0.13		& 0.67 & $\pm$ 0.07 & 0.74 & $\pm$ 0.06 & \bf 0.82 & $\pm$ 0.05 \tn
    \rowcolor{gray}
    C & 47.17 & $\pm$ 11.87 & 49.52 & $\pm$ 20.29 & 29.36 & $\pm$ 29.53 & 33.69 & $\pm$ 24.23 & 6.46 & $\pm$ 30.59 & −44.25 & $\pm$ 90.61	& 48.79 & $\pm$ 18.09 & 64.24 & $\pm$ 10.83 & \bf 77.46 & $\pm$ 7.31 \tn
    \hline
  \end{tabular}}
\end{table}

\begin{table}
 \caption{SCGM challenge results of the native MD-GRU with only CEL in comparison to the proposed GDL 0.5.
 Abbreviations of the metrics taken from \tabref{tab:GMchallengeResults}.
 }
 \label{tab:GMchallengeMDGRU}
 \resizebox{\textwidth}{!}{%
 \begin{tabular}{lrlrlrlrlrlrlrlrlrlrl}
 \rowcolor{gray}
    &\multicolumn{2}{c}{\bf DSC} &\multicolumn{2}{c}{\bf MD} &\multicolumn{2}{c}{\bf HD} &\multicolumn{2}{c}{\bf SHD} &\multicolumn{2}{c}{\bf SMD} &\multicolumn{2}{c}{\bf TPR} &\multicolumn{2}{c}{\bf TNR} &\multicolumn{2}{c}{\bf P} &\multicolumn{2}{c}{\bf J} &\multicolumn{2}{c}{\bf C} \tnhl
 \bf MD-GRU CEL & 0.87 & $\pm$ 0.03 & 0.30& $\pm$ 0.31& 2.14 & $\pm$ 1.20 & 0.85 & $\pm$ 0.36 & 0.40 & $\pm$ 0.20 & 93.93 & $\pm$ 3.85 &\bf 99.98 & $\pm$ 0.03 & 82.04 & $\pm$ 5.42 & 0.78 & $\pm$ 0.05 & 70.90 & $\pm$ 9.06 \tn
 \rowcolor{gray}
 \bf MD-GRU GDL 0.5 &\bf 0.90& $\pm$ 0.03 &\bf 0.21& $\pm$ 0.20 &\bf 1.85& $\pm$ 1.16 &\bf0.71& $\pm$ 0.28 &\bf 0.37& $\pm$ 0.17 &\bf 96.22& $\pm$ 2.69 &\bf99.98& $\pm$ 0.03 &\bf85.46& $\pm$ 4.96 &\bf0.82& $\pm$ 0.05 &\bf 77.46& $\pm$ 7.31 \tn
 \hline
 \end{tabular}}

\end{table}

\section{Conclusion}
\label{seq:conclusion}
We presented a new pipeline of acquisition and automatic segmentation of SC GM and WM.
The AMIRA sequence produces 8 channel images for different inversion times which the proposed deep learning approach with MD-GRU used for segmentation.
Using the 8 channels, tissue specific relaxation curves can be learned and used for GM-WM segmentation.

Comparing our segmentation results to the results of the ex-vivo high-reso\-lution dataset of Perone \etal \cite{perone_spinal_2018}, we show comparable accuracy for in-vivo data.
The acquired AMIRA dataset in scan-rescan fashion, with and without repositioning in the scanner, shows high reproducibility in terms of GM area RSD.
Thus we believe that the presented pipeline is a candidate for longitudinal clinical studies.
Further tests with patient data have to be conducted.

We added a generalized multi-label Dice loss to the cross entropy loss that MD-GRU uses.
We observed, that the segmentation performance was stable for a larger region of the weighting $\lambda$ between the two losses.
In a future work, we will study the effects of small $\lambda$s that correspond well with the logarithmical magnitudes of CEL.
Our proposed segmentation model outperforms the methods from the SC GM segmentation challenge.
Training the MD-GRU models directly on the 3D data might further improve the performance compared to slice-wise segmentation.

Given the small and fine structure of the GM, we like to point out, that the achieved results of the metrics are near optimal.
Higher resolutions of the imaging sequence will improve the accuracy more easily.

\vspace{5mm}
Acknowledgments: We thank Dr. Matthias Weigel, Prof. Dr. Oliver Bieri and Tanja Haas for the MR acquisitions with the AMIRA sequence.

\bibliographystyle{splncs04}
\bibliography{biblio.bib}

\end{document}